\documentclass[journal]{IEEEtran}
\ifCLASSINFOpdf
\usepackage[pdftex]{graphicx}
\else
\usepackage[dvips]{graphicx}
\fi
\usepackage{booktabs}
\usepackage{amsmath}
\usepackage{microtype}
\usepackage[caption=false]{subfig}
\begin{document}

\title{Towards Universal End-to-End Affect Recognition from Multilingual Speech by ConvNets}


\author{Dario~Bertero,~\IEEEmembership{Student~Member,~IEEE,}
        Onno~Kampman,~\IEEEmembership{Student~Member,~IEEE,}
        and~Pascale~Fung,~\IEEEmembership{Fellow,~IEEE}% <-this % stops a space
        \thanks{D. Bertero is with the Human Language Technology Center, Department of Electronic and Computer Engineering, The Hong Kong University of Science and Technology, Clear Water Bay, Hong Kong (e-mail: dbertero@connect.ust.hk).}% <-this % stops a space
        \thanks{O. Kampman is with the Human Language Technology Center, Department of Electronic and Computer Engineering, The Hong Kong University of Science and Technology, Clear Water Bay, Hong Kong (e-mail: opkampman@connect.ust.hk).}% <-this % stops a space
         \thanks{P. Fung is with the Human Language Technology Center, Department of Electronic and Computer Engineering, The Hong Kong University of Science and Technology, Clear Water Bay, Hong Kong (e-mail: pascale@ece.ust.hk).}% <-this % stops a space
\thanks{Manuscript received - }}


\markboth{Journal of \LaTeX\ Class Files,~Vol.~14, No.~8, August~2015}%
{Shell \MakeLowercase{\textit{et al.}}: Bare Demo of IEEEtran.cls for IEEE Journals}



\maketitle

\begin{abstract}
We propose an end-to-end affect recognition approach using a Convolutional Neural Network (CNN) that handles multiple languages, with applications to emotion and personality recognition from speech. We lay the foundation of a universal model that is trained on multiple languages at once. As affect is shared across all languages, we are able to leverage shared information between languages and improve the overall performance for each one. We obtained an average improvement of 12.8\% on emotion and 10.1\% on personality when compared with the same model trained on each
language only. It is end-to-end because we directly take narrow-band raw waveforms as input. This allows us to accept as input audio recorded from any source and to avoid the overhead and information loss of feature extraction. It outperforms a similar CNN using spectrograms as input by 12.8\% for emotion and 6.3\%
for personality, based on F-scores. Analysis of the network
parameters and layers activation shows that the network learns and extracts significant features in the first layer, in particular pitch, energy and contour variations. Subsequent convolutional
layers instead capture language-specific representations through
the analysis of supra-segmental features. Our model represents
an important step for the development of a fully universal
affect recognizer, able to recognize additional descriptors, such as stress, and for the future implementation into affective interactive systems.
\end{abstract}

\begin{IEEEkeywords}
universal affect recognition, speech, emotion, personality, end-to-end
\end{IEEEkeywords}

% \fi
\IEEEpeerreviewmaketitle

\section{Introduction}
% \IEEEPARstart{V}{irtual} agents, such as interactive chatbots and dialog systems, are  becoming more and more popular.

\IEEEPARstart{R}{ecognition} of human affect is a very important aspect of human communication. We not only convey messages by their literal meaning, but also by how they are expressed and other forms of non-verbal communication. This includes cues like tone of voice, gesture, facial expression, or even more subtle elements such as body temperature and heart rate~\cite{fung2016towards,li2015convolutional}. Many of the main affect characteristics are universal across different languages and cultures. This motivates the creation of universal models.
It is becoming increasingly important for machines to be able to recognize various forms of human affect. An affect recognition model should be universally applicable, not just in specific domains and languages. This will help us develop more advanced interactive systems~\cite{fung2016zara,winata2017nora} that are able to detect and use affect, in addition to standard ASR and NLP techniques, to provide advanced services such as personality analysis, counseling, education, medical, or commercial services. We focus on affect recognition from audio and speech in this work through two universal affect characteristics retrievable from speech, namely emotion and personality. 

In the field of emotion detection, there is no general agreement on the number of basic emotion descriptors~\cite{ortony1990s}. It ranges from three (\emph{Anger}, \emph{Happiness} and \emph{Sadness}, with the eventual inclusion of the \emph{Neutral} class) to up to 20 for some commercial services. Each available corpus includes a different set of emotions. These emotions are often projected onto a plane formed by two main axes: Valence and Arousal~\cite{lang1993looking}. This way the classification task is reduced to the prediction of these two scores and the identification of a point on the plane. This greatly simplifies the classification process and training procedure, but it is less natural for humans to understand and interpret the meaning of Valence and Arousal compared to discrete emotions labels. Furthermore, it poses difficulties and uncertainties in the annotation process. Various emotion types are usually obtained through clustering the plane. For these reasons, and to provide more detailed analysis on each emotion, we decided to perform classification on discrete emotion values in our work described in this paper.

For personality recognition a standard set of descriptors are five personality \textit{traits} from the Big Five model~\cite{digman1990personality}. Traits are patterns in thought and behavior. An individual scores between 0 (low) and 1 (high) for each trait. Thus, an individual's personality is represented by a 5-dimensional vector of scores for the following traits:

\begin{itemize}
\item \emph{Extraversion} refers to assertiveness and energy level. Low scorers in this trait are more reserved and calm. 
\item \emph{Agreeableness} refers to cooperative and considerate behavior. Low scorers in this trait are less interested in social harmony with those around them.
\item \emph{Conscientiousness} refers to behavioral and cognitive self-control. Low scorers in this traits are typically seen as more irresponsible and disorganized.
\item \emph{Neuroticism} refers to a person's range of emotions and his/her control over these emotions. Low scorers are often more chaotic and anxious.
\item \emph{Openness to Experience} refers to creativity and adventurousness. Low scorers in this trait are typically more conservative and less curious.
\end{itemize}

We are particularly interested in whether affect is language-dependent and whether we can build a universal affect recognition system. Some researchers found that emotions vary only a little from one language or culture to the other~\cite{izard2013human}. The Big Five traits of personality have been demonstrated to be quite robust over different geographic locations~\cite{schmitt2007geographic}. Our previous work has shown that the manifestation of at least some affect, such as stress, is more gender-dependent than language-dependent~\cite{zuo2011cross}. It is interesting for us to investigate what features of speech, if any, are language independent. Finding commonality in the speech of different languages has shown to be effective for multilingual acoustic modeling, where it shares part of their phone set at the model level, the state-level, or the acoustic feature level~\cite{li2011asymmetric,schultz2001language,fung2008multilingual}. Multilingual models are also beneficial with the scarcity of training data in one or multiple languages. For our task on affect recognition, there is not a huge amount of human-labeled data available in any language. We postulate that a universal end-to-end model shared between different language samples may improve the performance on each individual language.

Another objective of our work is to explore machine learning methods that can extract features automatically from raw waveforms, without explicit human design. We are motivated by the fact that the human auditory system is capable of processing audio in different languages without any morphological changes. In addition, previous work on multilingual speech recognition has shown that there are stronger responses in certain groups of the spectral frequencies to phonetic sounds in certain languages. The class of deep learning algorithms called Convolutional Neural Networks (CNN) has shown to be astute in automatic feature extractions in both the image and speech domains~\cite{krizhevsky2012imagenet,abdel2014convolutional}. We aim to investigate end-to-end CNN models for affect recognition.  An important objective of using automatic feature extraction combined with classification is to bypass the time delay in extracting features. A system based on narrow-band raw waveforms would allow us to avoid any corpus-dependent and language-dependent feature engineering step, would be applicable to any sort of spoken input signal, such as phone calls, and would require less memory and pre-processing overhead.

This work is a significant extension of our earlier attempts to detect speech emotions from narrow-band speech raw waveforms~\cite{berteroemnlp,bertero2017icassp} to include personality analysis and a multilingual approach. In this paper we significantly revise the model and experimental setup from our previous works and experiments on more datasets in different languages. We do not only limit the application to the emotion detection problem, but we also show the effectiveness on the more difficult personality detection from speech, again in a multilingual setting. We then provide more insights about what the model is actually trying to learn, and how it generalizes across languages. 

\section{Related work}
\label{related_work}

\subsection{Multilingual approaches for speech and language}
Multilingual approaches from speech and language first appeared in the 1990s with statistical models. These models have been found to help improve the performance for those languages with limited resources, through taking advantage of the similarities among different languages and by borrowing from resource-rich languages~\cite{fung2008multilingual}. 

Since we are interested in deep learning methods that can automatically extract features, we also look at recent multilingual neural models that have been proposed in speech processing, including speech recognition~\cite{huang2013cross,heigold2013multilingual,ghoshal2013multilingual,dey2014acoustic,thomas2013deep,mohan2015multi,knill2013investigation,grezl2014adaptation}. Neural network architectures and training procedures were specifically designed to handle multilingual input and take advantage of multiple languages combined to improve the recognition performance on each of them.

For Automatic Speech Recognition (ASR), \cite{huang2013cross,heigold2013multilingual} used a multilayer DNN with an array of language-specific final layers to share the acoustic features across different languages. \cite{ghoshal2013multilingual,dey2014acoustic} instead used a similar DNN with a single final layer fine-tuned on different languages, while \cite{thomas2013deep} applied a progressive layer by layer training first from a multilingual corpus and then from specific languages. Other techniques were used to adapt the final layers such as low-rank factorization of parameter matrices \cite{mohan2015multi} or bottleneck layers and extra features \cite{knill2013investigation,grezl2014adaptation}. 

\subsection{Emotion recognition from speech}
Previously speech emotion detection was performed through the extraction of many features from the audio sample which are then fed into a supervised classifier \cite{eyben2009openear}. A standard set of features included speech features such as MFCC, psycholinguistic features \cite{cabrera2007psysound3} and other low-level audio features such as pitch, zero-crossing rate, energy and many others~\cite{eyben2009openear}. They were extracted from small audio frames, typically of around $25\,\text{ms}$, and then combined together to represent the utterance to analyze. This combination was performed either through many statistical functionals such as mean, standard deviation, skewness, kurtosis, etc.~\cite{stuhlsatz2011deep}, or directly through the classifier~\cite{han2014speech}. The classifier choice ranged from basic supervised classifiers such as SVM~\cite{eyben2009openear,zhang2011unsupervised} and decision trees~\cite{lee2011emotion}, to more complex deep learning structures such as DNN~\cite{zhang2017multi}, CNN \cite{aldeneh2017using}, ELM~\cite{han2014speech} or LSTM in the case of continuous emotion detection~\cite{mirsamadi2017automatic}.
Most of the analysis was performed on the valence-arousal plane \cite{eyben2009openear}, often grouping multiple discrete emotions as high/low valence and high/low arousal \cite{schuller2010cross,zhang2011unsupervised}.

All those feature sets were often collected and provided in various shared task \cite{schuller2009interspeech,schuller2010interspeech,schuller2013interspeech}, and used as standard feature sets for affect recognition thereafter. Others have applied more complex feature engineering and feature selection~\cite{liu2010novel}, but these processes are often time consuming, add overhead latency or be database dependent. Departing from the traditional feature engineering approach \cite{schmidt2011learning,mao2014learning} used deep learning models to perform automatic feature extractions from the audio represented as a spectrogram. In these works the spectrograms are described as the ``raw audio signal''. However, we note that the spectrogram itself, though more limited in scope, is already a feature extraction step where each audio frame is associated with its FFT coefficients, thus it is not the ``raw audio signal'' as purported. We are interested in investigating whether CNNs can extract features and classify them correctly directly from the time-domain raw waveform.

Extending the analysis to the field of multilingual and cross-domain emotion and personality recognition, most works applied traditional feature-engineering with eventual speaker-normalization \cite{schuller2010cross, schuller2011using}. \cite{sagha2016cross} tried to solve the problem through the extraction of a shared feature representation using Kernel Canonical Correlation Analysis \cite{hardoon2004canonical}, while \cite{deng2013sparse, deng2014introducing} obtained the shared feature representation training an autoencoder. \cite{zhang2011unsupervised} managed to increase the classification accuracy over various corpora by using unlabeled data.

With the exception of \cite{trigeorgis2016adieu} and our preliminary work \cite{berteroemnlp,bertero2017icassp}, no other work to our knowledge have ever tried to classify paralinguistic traits using raw waveforms as input directly. Even  \cite{trigeorgis2016adieu} analyzed only a very limited dataset, and only on the valence-arousal dimensions, providing very limited insights on the proposed network architecture.

\subsection{Personality recognition from speech}
The field of Personality Computing is quite young, but some work has been published on recognizing Big Five personality traits from the non-verbal part of speech~\cite{vinciarelli2014survey}. Just as in emotion detection, most of the work has focused on extracting low-level prosodic features and statistical functions thereof, using a standard classification algorithm such as SVM or Random Forest to determine whether the subject scores \textit{above} or \textit{below} the median for each of the five traits~\cite{mairesse2007using, pianesi2008multimodal, polzehl2010automatically}.  
%Although dividing the continuous scale per trait into two classes (typically above or below average) is useful for comparing classification results, these classes have little psychological meaning.

The \textit{Interspeech 2012 Speaker Trait Challenge}~\cite{schuller2012interspeech} was the first comprehensive effort to compare different approaches to the problem, by benchmarking on the same test dataset. Popular prosodic features are statistics of pitch, energy, first two formants, length of voided and unvoiced segments, as well as Mel Frequency Cepstral Coefficients (MFCC). The winner of the competition extracted thousands of spectral features before doing a feature selection process~\cite{ivanov2012modulation}, a method that was very common~\cite{pohjalainen2012feature}. These features were then mapped to the five traits through SVM classifiers. Classification accuracies from this challenge are between 60 and 75 percent, depending on the trait. Although many different approaches and machine learning algorithms were tried, none of them clearly outperformed the others. Also, due to the limited number of subjects, the results from these works are often statistically unreliable and could be heavily corpus-dependent.

Related work found that the Extraversion and Conscientiousness traits were easiest to classify, and Openness to Experience the hardest~\cite{mohammadi2010voice, polzehl2010automatically, zhang2017social}.

The ChaLearn 2016 shared task challenge released a large corpus of 10,000 extracts from YouTube video blogs~\cite{ponce2016chalearn}. Each clip was labeled with continuous Big Five labels. Each of the participants of the shared task used audio as well as video, and it is not possible to directly look at recognition performance from just audio. This workshop is still interesting for two reasons. The corpus provided is, to our knowledge, the biggest open-domain personality corpus, and the best performing teams used neural network techniques. However, although teams inserted video directly into a neural network, they still extracted traditional audio features (zero crossing rate, energy, spectral features, MFCCs) that were then fed into the neural network~\cite{subramaniam2016bi, gurpinar2016multimodal, zhang2016deep}. A deep neural network should however be able to extract such features itself (if relevant for the task). An exception was~\cite{guccluturk2016deep}, but they used a neural network specialized for image processing and computer vision, and did not adapt it to audio input. The team with the best performance in the challenge extracted openSMILE~\cite{eyben2009openear} acoustic features as used in the INTERSPEECH 2013 challenge baseline set~\cite{schuller2013interspeech}, which they linearly mapped to the Big Five traits. They did not publish their work. The challenge was aimed at the computer vision community (many only used facial features), thus although many teams analyzed their approaches to vision, not many looked into detail what their deep learning network was learning regarding audio input.

\section{Methodology}
\label{methodology}
In this paper, we propose a method for automatically recognizing emotion and personality from speech in different languages, without the need for feature extraction upfront. We propose to achieve this with a multi-layer Convolutional Neural Network (CNN) framework, trained end-to-end from raw waveforms. We train models in monolingual and multilingual settings. We compare our model with a similar CNN model that takes spectrogram representations as input.

\subsection{Preprocessing}
We are interested in recognizing affect from a given input audio sample. The very first processing step is to downscale the input sample to a uniform sampling rate. We choose narrowband speech at $8\,\text{kHz}$ for our work. There are two main reasons for this choice. The first is to analyze how the system would work under the worst possible conditions, for example to detect emotion over a phone call. The second one is to reduce the eventual transmission time and memory requirements when the speech has to be sent over internet or has to be stored and processed in an embedded system, it also reduces the number of network parameters.

An aspect often overlooked while designing models for affect recognition is the input volume. Features such as relative energy within or across frames are important components of affect, as sudden changes may signal high arousal emotions like \emph{anger}. However absolute energy over the entire sample is not useful, as it mainly depends on the volume the sample was recorded at. The absolute energy level may cause severe overfitting to the model. This is often evident with emotions like \emph{anger} or \emph{sadness}, where sometimes the model only learns to distinguish the classes through the amplitude level ignoring the rest of the features. Different language corpora, especially when consisting of spontaneous speech, may contain samples recorded at varying input volumes. The position of the speaker with respect of the microphone may also differ each time. All these aspects cannot be determined a priori. Volume normalization techniques, such as peak or RMS normalization, can be applied but they are not fully suitable for our task for the following reasons: 1) Peak normalization would suffer from isolated peaks which are often not representative of the whole sample; 2) RMS normalization instead would be sensibly different depending on the amount of silence in the audio sample, which is not always related to affect (potentially due either to a low speaking rate, pauses in the recording or the microphone kept open).    

Starting from the assumption that affect does not change depending on the recording volume, during the training phase, but not during the evaluation, we randomized uniformly the amplitude (volume) of the input audio sample through an exponential random coefficient $\alpha$ at each training iteration, where $\alpha$ is equal to
\begin{equation}
\label{volnorm}
	\alpha = 10^{\text{U}(-a, a)}
\end{equation}
where $\text{U}(-a, a)$ is a uniform random variable, and $a$ a hyperparameter. We applied this pre-processing instead of normalizing the volume to a fixed value in order to increase the robustness of the system. A uniform random variable over a wide range of value (compared to a normal distribution) not only as said before helps reducing the overfitting related to the energy component, but also strongly augments the training set size. 

\begin{figure}[t!]
  \centering
  \includegraphics[width=0.8\columnwidth]{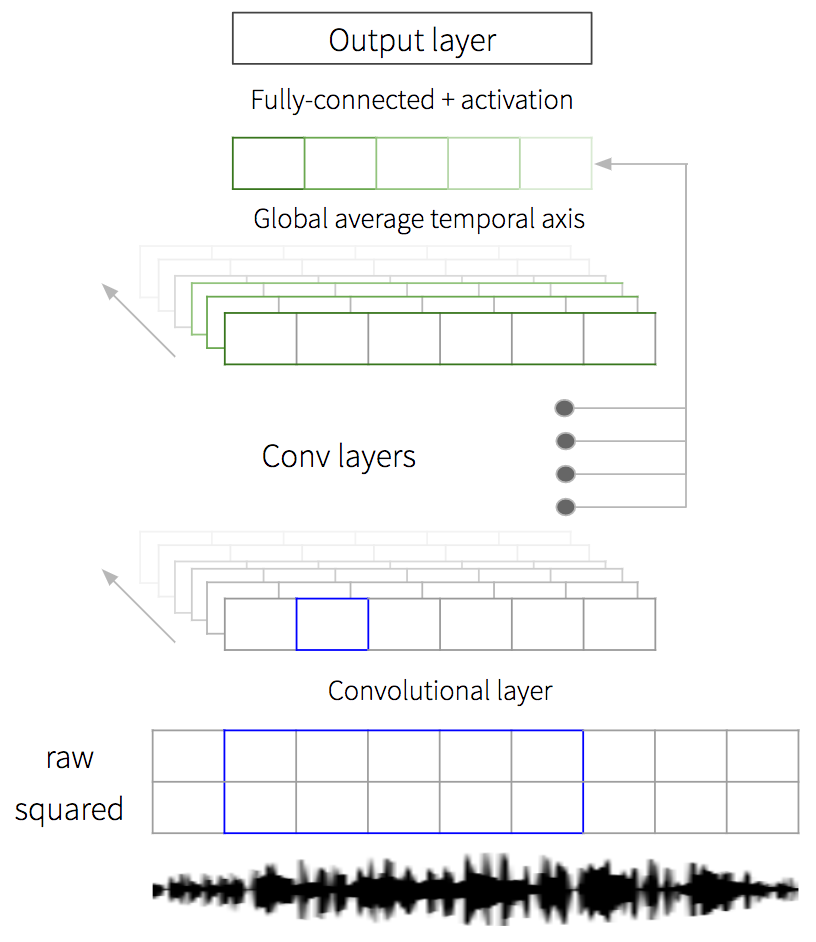}
  \caption{Convolutional Neural Network architecture for emotion and personality recognition from raw waveforms. The output consists of either the four emotion classes analyzed or the Big Five personality trait scores.}
  \label{model}
\end{figure}

\subsection{CNN for feature extraction}
The aim of our work is, given an input utterance or speech segment in the form of raw waveforms, to determine the overall affect state expressed. The Convolutional Neural Network is an ideal architecture for this task as it is able in sequence to learn and perform feature extraction from short overlapping windows regardless of the overall sample length, analyze the variation over neighboring regions of different sizes, and combine all these contributions into an overall vector for the entire audio sample. CNN are typically employed with great success in image recognition task. In acoustic analyses, audio samples can be regarded as 1-dimensional ``images''. Each component does not represent a pixel but the value of the acoustic waveform, and different input channels may include different signal transformations. Ideally our model should also learn to internally extract features and process audio consisting of different speaker characteristics, such as gender and age, different languages and different input volumes without any prior normalization. 

This process is similar to that applied by traditional feature-based methods. In these methods a feature extraction tool, such as openSMILE~\cite{eyben2009openear}, is used to extract a series of features (typically MFCC, pitch, zero-crossing rate and energy) from the audio sample divided into small frames. Frame-based features are then merged together with a series of higher-level descriptors (such as mean, standard deviation, skewness, kurtosis, etc.). However, the features and the high-level descriptors are not statically defined a priori, but are learnt by the network. We expect that low-level features would be mostly extracted by the first layer, and high-level descriptors by the higher layers \cite{golik2015convolutional}. The network would also presumably learn to automatically filter the ones less useful, concentrate more on those more useful and eventually extract some other different features which were not usually applied in affect recognition before.

\subsection{CNN model description}
Our CNN consists of a stack of convolutional layers of different sizes. It is followed by a global average pooling operation on the output of each layer, a weighted average combination of all these vectors, a fully connected layer and final activation layer (softmax for emotion and sigmoid for personality). The specific role of each layer will be described in detail in Section~\ref{discussion}. A model diagram is shown in Figure~\ref{model}.

The CNN receives as input a raw sample waveform $\mathbf{x}$ of narrow-band speech, sampled at $8\,\text{kHz}$, of arbitrary length. We split the input signal into two feature channels as input for the CNN. The first one is the raw waveform as-is, the second one is the signal with squared amplitude. The second channel is mostly aimed at capturing the energy component of the signal and learn an implicit normalization.

The two input signal components are then directly fed into a first convolutional layer:
\begin{equation}
  \mathbf{x}^{(1)}_{i} = f(\mathbf{W}^{(1)} \mathbf{x}_{[i,i+v]} + \mathbf{b}^{(1)})
\end{equation} 
where $v$ is the convolution window size and $f$ a non-linear function. In this first layer we use a window size of 200, which at $8\,\text{kHz}$ sampling rate corresponds to $25\,\text{ms}$, and a stride of 100, which corresponds to around $12\,\text{ms}$. The output size $\text{dim}(\mathbf{x}^{(1)}_{i})$ (the number of filters trained) from each window is set to 512. This first layer acts as a low-level feature extractor, or a customized filterbank learnt over the corpus during training. It ideally replaces the discrete features extraction step or the FFT computation of the spectrogram window. The window length of $25\,\text{ms}$ is a common choice for the feature extraction step, as shown in previous works using feature-based or spectrogram based CNNs~\cite{schuller2009interspeech,aldeneh2017using}. 

It is then followed by several higher convolutional layers, of the same number of filters. Their convolution window size and stride is set to capture increasingly larger time spans. 
The subsequent convolutions are aimed at combining the features and capturing information at the suprasegmental level, such as phonemes, syllables and words, as well as looking at the difference between contiguous frames.

Since our intent is to capture globally expressed emotion and personality characteristics from the entire audio frame, the contributions of the convolutional layer outputs must be combined together. This is done by a global average pooling operation over the output vectors:
\begin{equation}
  \mathbf{x}_j^{\text{AP}} = \dfrac{\sum_i (x_{i,j})}{L_i}
\end{equation}
where $i$ is the window index, $L_i$ the number of output windows from the convolution layer, and $j$ the feature vector index within each convolution window. The average pooling is performed over the output vectors of each layer instead of just the final one. This would combine the contributions of both the segmental and suprasegmental features at different temporal granularities for the final emotion and personality decision. The output average-pooling vectors for each layer are then combined through a weighted-average layer where the weights are parameters of the model:
\begin{equation}
\label{wavg}
	\mathbf{x}^{\text{OUT}} = f\left(\sum_l \mathbf{W}^{\text{OUT}}_l\,\mathbf{x}^{\text{AP}}_l + \mathbf{b}_{\text{OUT}}\right) 
\end{equation}
where $l$ is the layer index, and $f$ again the non-linear function.

We decided to use average pooling instead of the more common max-pooling. This choice yielded higher results on the development sets. It is also meaningful as the objective of our work is to detect the overall affect of an entire utterance or a speech passage. The global average pooling can be seen as merging together all the intermediate affect results~\cite{}. It sums and accumulates the contributions among all the speech segments considered, instead of just selecting a few salient instances. We empirically noticed that applying max-pooling, even side-by-side with average pooling, makes the network overfit the training data more easily.

After obtaining the audio-frame overall vector $\mathbf{x}^{\text{OUT}}$ by weighted-average of each convolutional layer output (Eq. \ref{wavg}), we then feed it through a fully connected layer, followed by a final softmax/sigmoid layer. This last layer performs the final classification/regression operation and outputs the probability of the sample to belong to each emotion class analyzed as well as the personality trait scores.

In each of the intermediate layers the exponential linear activation function is used as non-linearity~\cite{clevert2015fast}, as it performed better on the development set compared to other popular choices such as the hyperbolic tangent ($\text{tanh}$) or the rectified linear function (ReLU).

\subsection{Multilingual adaptation}
Our CNN is already designed to handle a multilingual setting taking advantage of data in different languages. The duty of the first layer is to learn and extract low-level features common across all languages, such as filterbank features, pitch and energy. More data can improve this step's performance. The subsequent layers are instead delegated to supra-segmental features, some of which are specific to languages or groups of languages. The application of a large layer size, 512 in our architecture, also allows the network to better learn these language-specific features and language acoustic models.

Although the model is already adequate to learn affect from multiple languages, further language-specific adaptation is desirable. After the initial training on the full dataset, we retrain the final layers after the average pooling on a single language data. This adaptation, or fine-tuning step, operates by weighting  differently the extracted features of each layer, in order to adapt to each specific language analyzed. It is here where different affect states are communicated that can be dependent on language.

\begin{table*}[h]
  \centering
  \caption{Number of utterances of each class, and total number, in the emotion corpora considered. In parenthesis the division among training and test set.}
  \scalebox{0.85}{
  \begin{tabular}{l|cccc|c}
    & \multicolumn{3}{c}{Speakers} & \\
    Language & Anger & Sadness & Happiness & Anxiety & Total utterances \\
    \hline
    English & 1202 (1092/110) & 1246 (1115/131) & 2128 (1933/195) & 952 (865/87) & 5528 \\
    Estonian \cite{altrov2013estonian} & 306 (275/31) & 249 (224/25) & 271 (243/28) & - & 826 \\
    German \cite{burkhardt2005database} & 127 (102/25) & 62 (54/8) & 71 (65/6) & 68 (62/6) & 328 \\
    Spanish \cite{hozjan2002interface} & 725 (652/73) & 731 (657/74) & 732 (658/74) & 735 (661/74) & 2923 \\
    Italian \cite{costantini2014emovo} & 84 (56/28) & 84 (56/28) & 84 (56/28) & 84 (56/28) & 336 \\  
    Serbian \cite{jovicic2004serbian} & 366 (244/122) & 366 (244/122) & 366 (244/122) & 366 (244/122) & 1464 \\
    \hline
    Total & 2810 & 2738 & 3652 & 2205 & 11405 \\
  \end{tabular}
  }
  \label{emotion_speakers}
\end{table*}

\subsection{Spectrogram CNN}
Until recently the idea of using the raw representation of a signal often refers to a spectrogram presented as an image to a CNN~\cite{schmidt2011learning,mao2014learning}. As a comparison baseline, we propose a similar CNN that takes the spectrogram representation as input.

The spectrogram CNN is very similar to the one used for raw waveforms. A spectrogram representation is first extracted from a raw input waveform, again sampled at $8\,\text{kHz}$. This is done through a Tukey window of $25\,\text{ms}$, with an FFT-size of 256, and yields a series of 129 power spectral features for each window. This operation replaces the feature extraction done by the first convolution layer of the raw waveform network. The subsequent layers are the same as in the raw waveform network, including several convolutional layers, global average pooling for each layer, weighted-average, fully connected and activation layers.

\section{Emotion Recognition Experiments}
\label{emotion_experiments}
\begin{table*}[t!]
  \centering
   \caption{Results (percentage) on multilingual task for the four emotions analyzed (anger, sadness, happiness, anxiety). P stands for precision, R for recall, and F1 for F-score (or F1 measure).}
  \scalebox{0.85}{
  \begin{tabular}{l|ccc|ccc|ccc|ccc|ccc|ccc}
    & \multicolumn{3}{c}{English} & \multicolumn{3}{c}{Estonian} &  \multicolumn{3}{c}{German} &  \multicolumn{3}{c}{Spanish} &  \multicolumn{3}{c}{Italian} & \multicolumn{3}{c}{Serbian} \\
    Method & P & R & F1 & P & R & F1 & P & R & F1 & P & R & F1 & P & R & F1 & P & R & F1 \\
    \hline
    \multicolumn{19}{c}{\textbf{Anger}}\\
    \hline
  Single lang. spec & 0.0 & 0.0 & 0.0 & 45.1 & 74.2 & 56.1 & 82.6 & 76.0 & 79.2 & 92.8 & 87.7 & 90.1 & 93.3 & 50.0 & 65.1 & 56.7 & 45.1 & 50.2 \\
  Multilingual spec & 30.1 & 25.5 & 27.6 & 56.7 & 54.8 & 55.7 & 77.3 & 68.0 & 72.3 & 76.3 & 83.6 & 79.7 & 48.4 & 53.6 & 50.8 & 67.0 & 50.0 & 57.3 \\
  Single lang. raw & 44.6 & 40.9 & 42.7 & 40.0 & 96.8 & 56.6 & 76.9 & 80.0 & 78.4 & 86.5 & 87.7 & 87.1 & 61.9 & 46.4 & 53.1 & 47.8 & 73.0 & 57.8 \\
  Multilingual raw & 49.6 & 54.5 & 51.9 & 54.5 & 77.4 & 64.0 & 82.6 & 76.0 & 79.2 & 93.2 & 94.5 & 93.9 & 68.4 & 46.4 & 55.3 & 58.2 & 87.7 & 69.9 \\
  Fine-tuned raw & 46.2 & 54.5 & 50.0 & 56.4 & 71.0 & 62.3 & 83.3 & 80.0 & 81.6 & 95.8 & 94.5 & 95.2 & 77.8 & 50.0 & 60.9 & 67.2 & 70.5 & 68.8 \\
    \hline
    \multicolumn{19}{c}{\textbf{Sadness}}\\
    \hline
  Single lang. spec & 62.4 & 44.3 & 51.8 & 75.0 & 36.0 & 48.6 & 100.0 & 100.0 & 100.0 &  95.9 & 95.9 & 95.9 & 73.3 & 39.3 & 51.2 & 95.7 & 90.2 & 92.8 \\
  Multilingual spec & 62.0 & 37.4 & 46.7 & 50.0 & 60.0 & 54.5 & 100.0 & 87.5 & 93.3 & 93.4 & 95.9 & 94.7 & 85.2 & 82.1 & 83.6 & 80.1 & 95.9 & 87.3 \\
  Single lang. raw & 60.8 & 47.3 & 53.2 & 0.0 & 0.0 & 0.0 & 80.0 & 100.0 & 88.9 & 93.0 & 89.2 & 91.0 & 66.7 & 71.4 & 69.0 & 91.7 & 90.2 & 90.9 \\
  Multilingual raw & 64.2 & 65.6 & 64.9 & 52.2 & 48.0 & 50.0 & 100.0 & 100.0 & 100.0 & 96.1 & 100.0 & 98.0 & 62.2 & 82.1 & 70.8 & 88.3 & 99.2 & 93.4 \\
  Fine-tuned raw & 64.2 & 60.3 & 62.2 & 57.1 & 48.0 & 52.2 & 100.0 & 100.0 & 100.0 & 96.1 & 100.0 & 98.0 & 76.0 & 67.9 & 71.7 & 91.0 & 99.2 & 94.9 \\

  \hline
  \multicolumn{19}{c}{\textbf{Happiness}}\\
  \hline
  Single lang. spec & 42.3 & 93.3 & 58.2 & 57.1 & 42.9 & 49.0 & 25.0 & 50.0 & 33.3 & 86.1 & 91.9 & 88.9 & 71.9 & 82.1 & 76.7 & 44.1 & 68.0 & 53.5 \\
  Multilingual spec & 43.2 & 66.7 & 52.4 & 58.3 & 50.0 & 53.8 & 15.4 & 33.3 & 21.1 & 80.3 & 71.6 & 75.7 & 26.1 & 21.4 & 23.5 & 47.0 & 64.8 & 54.5 \\
  Single language raw & 52.6 & 77.4 & 62.7 & 22.2 & 7.1 & 10.8 & 0.0 & 0.0 & 0.0 & 81.3 & 87.8 & 84.4 & 58.6 & 60.7 & 59.6 & 51.5 & 42.6 & 46.6 \\
  Multilingual raw & 61.5 & 63.1 & 62.3 & 58.8 & 35.7 & 44.4 & 28.6 & 33.3 & 30.8 & 89.3 & 90.5 & 89.9 & 68.2 & 53.8 & 60.0 & 61.2 & 51.6 & 56.0 \\
  Fine-tuned raw & 63.2 & 62.6 & 62.9 & 54.2 & 46.4 & 50.0 & 20.0 & 16.7 & 18.2 & 89.6 & 93.2 & 91.4 & 73.9 & 60.7 & 66.7 & 55.7 & 59.8 & 57.7 \\
    \hline
    \multicolumn{19}{c}{\textbf{Anxiety}}\\
    \hline
    Single lang. spec & 0.0 & 0.0 & 0.0 & - & - & - & 66.7 & 28.6 & 40.0 & 91.8 & 90.5 & 91.2 & 46.0 & 82.1 & 59.0 & 69.3 & 50.0 & 58.1 \\
  Multilingual spec & 14.0 & 8.0 & 10.2 & - & - & - & 50.0 & 28.6 & 36.4 & 87.7 & 86.5 & 87.1 & 61.3 & 67.9 & 64.4 & 72.3 & 49.2 & 58.5 \\
  Single lang. raw  & 36.4 & 13.8 & 20.0 & - & - & - & 83.3 & 71.4 & 76.9 & 90.0 & 85.1 & 87.5 & 28.1 & 32.1 & 30.0 & 69.1 & 45.9 & 55.2 \\
  Multilingual raw  & 23.5 & 18.4 & 20.6 & - & - & - & 87.5 & 100.0 & 93.3 & 98.6 & 91.9 & 95.1 & 64.7 & 78.6 & 71.0 & 84.4 & 44.3 & 58.1 \\
  Fine-tuned raw & 24.7 & 21.8 & 23.2 & - & - & - & 77.8 & 100.0 & 87.5 & 98.6 & 91.9 & 95.1 & 54.3 & 89.3 & 67.6 & 80.2 & 63.1 & 70.6 \\
    \hline
    \multicolumn{19}{c}{\textbf{Average}}\\
    \hline
  Single lang. spec & 26.2 & 44.4 & 27.5 & 59.1 & 51.0 & 51.2 & 68.6 & 63.7 & 63.1 & 91.6 & 91.5 & 91.5 & 71.1 & 63.4 & 63.0 & 66.5 & 63.3 & 63.7 \\
  Multilingual spec & 37.3 & 34.4 & 34.2 & 43.0 & 54.9 & 54.7 & 60.1 & 54.4 & 55.8 & 84.4 & 84.4 & 84.3 & 55.3 & 56.3 & 55.6 & 66.6 & 65.0 & 64.4 \\
  Single lang. raw  & 48.6 & 44.9 & 44.7 & 20.7 & 34.6 & 22.4 & 60.1 & 62.9 & 61.1 & 87.7 & 87.4 & 87.5 & 53.8 & 52.7 & 52.9 & 65.0 & 62.9 & 62.6 \\
  Multilingual raw  & 49.7 & 50.4 & \textbf{49.9} & 55.2 & 53.7 & 52.8 & 74.7 & 77.3 & \textbf{75.8} & 94.3 & 94.2 & 94.2 & 65.9 & 65.2 & 64.3 & 73.0 & 70.7 & 69.4 \\
  Fine-tuned raw & 48.9 & 49.8 & 49.6 & 55.9 & 55.1 & \textbf{54.8} & 70.3 &  74.2 & 71.8 & 95.0 & 94.9 & \textbf{94.9} & 70.5 & 67.0 & \textbf{66.7} & 73.5 & 73.2 & \textbf{73.0}\\
  \end{tabular}
  }
  \label{res_emotion}
\end{table*}

\subsection{Corpora}
In our experiments we make use of two set of corpora: a multi-domain English corpus with crowdsourced labels, and a set of smaller corpora of acted emotions in various languages. A summary of the number of utterances of each corpus is shown in Table~\ref{emotion_speakers}.

The English corpus is made of data we collected and annotated in multiple phases over time \cite{berteroemnlp,bertero2016towards}. We collected thousands of utterances and short speeches from different sources including monologues (TED talks, YouTube vloggers) and dialogues (TV shows, debates). In the case of TV shows, individual utterances were segmented from the audio track using the subtitles timestamps as references. The monologues instead were cut into segments of around $10-15\,\text{s}$, using silences as references.  We then labeled them with several emotion descriptors, using student helpers and through Amazon Mechanical Turk. Each audio clip was annotated by a minimum of one annotator (in the case of the student helpers, previously instructed on the task) to a maximum of five annotators. We took the label selected by the majority of the annotators, discarding the sample in case of disagreement. In this work we only consider the subset of utterances classified as \emph{anger}, \emph{sadness}, \emph{happiness} and \emph{anxiety}. We also annotated the data with other emotions labels. However, some of them were not present in all languages. Others contained a number of samples too limited for training.

To train a universal multilingual model and evaluate the performance of our classifier on different languages, we used several corpora listed below. Compared to the English database, they contain a limited number of speakers who were actors that generated each emotion in a studio setting. Source sampling rate was usually $16\,\text{kHz}$ or higher.
\begin{itemize}
\item \textbf{Estonian - Estonian Emotional Speech Corpus} \cite{altrov2013estonian}: the corpus consists of 1234 Estonian utterances. They are generated by a single actress in four emotions: Anger, Joy, Sadness and Neutral.
\item \textbf{German - Berlin EmoDB} \cite{burkhardt2005database}: this database consists of 535 German utterances. A total of 5 short and 5 long utterances were generated by 5 actors and 5 actresses in 7 emotions: Anger, Neutral, Fear, Boredom, Happiness, Sadness and Disgust (not all the actors read all the sentences for each emotion).
\item \textbf{Spanish - INTERFACE Emotional Speech Syntesis Database} \cite{hozjan2002interface}: this database includes around 150 items (phonemes, words, short, long sentences and a longer ~30 s passage) in Spanish language. Each item is spoken by a male and a female actors in several emotions: Anger, Sadness, Joy, Surprise, Disgust, Fear and Neutral. For the purpose of our work we discarded the phonemes and the individual words.
\item \textbf{Italian - EMOVO} \cite{costantini2014emovo}: emotion corpus in Italian. It includes 6 actors (3 males and 3 females), each acting 14 sentences into 7 different emotions: Anger, Neutral, Disgust, Joy, Fear, Surprise, Sadness.
\item \textbf{Serbian - Serbian Emotional Speech Database} \cite{jovicic2004serbian}: made of 3 actors and 3 actresses and a total of 2694 utterances, including one longer passage but excluding the isolated word part of the database. It includes five emotions: anger, happiness, fear, sadness and neutral.
\end{itemize}

In our work we analyze a subset of emotion labels common to most of the corpora: namely \emph{anger}, \emph{sadness}, \emph{happiness} and \emph{anxiety}. As each database is made of slightly different emotions or denominations, we take \emph{fear} as \emph{anxiety} and \emph{joy} as \emph{happiness}.

\begin{table*}[t!]
  \label{tab:res_personality}
  \centering
   \caption{Results (percentage) on multilingual task for Big Five personality traits analyzed. Mean Absolute Error (MAE), Accuracy (A), Precision (P), Recall (R), and F-score (F1) are shown.}
   \scalebox{0.85}{
  \begin{tabular}{l|ccccc|ccccc|ccccc}
  
      					&\multicolumn{5}{c}{English} 		&\multicolumn{5}{c}{Mandarin} 			&\multicolumn{5}{c}{French} 	\\
    Method 				&MAE &A &P &R &F1 					&MAE &A &P &R &F1 						&MAE &A &P &R &F1 				\\
    \midrule
    \multicolumn{16}{c}{\textbf{Extraversion}}																						\\
    \midrule
%     Single lang. spectrogram     &.107 && 66.7 & 69.7 & 68.2 &.057 &46.7 &- &0 &- &.094 &68.9 &70.9 &76.1 &73.5 \\
%     Single lang. raw audio       &.107 &66.0 &63.0 &76.6 &69.1 &.043 &73.5 &67.1 &82.8 &74.1 &.085 &74.1 &74.0 &83.5 &78.4 \\
%     Single lang. raw3    &.107 &66.0 &63.0 &76.6 &69.1 	&.043 &73.5 &67.1 &82.8 &74.1 		&.085  &74.1 &74.0 &83.5 &78.4 		\\    
%     Single lang raw   &.1109 &62.4 &65.5 &57.9 &61.4 		&.0823 &56.7 &66.7 &3.8 &7.1 		&.1129 &56.9 &72.0 &39.6 &51.1 		\\    
	Single lang. spec	&.1093 &63.7 &67.4 &57.9 &62.3		&.0449 &70.8 &68.0 &64.2 &66.0 		&.1043 &69.4 &73.3 &72.5 &72.9 		\\
    Single lang. raw   	&.1141 &61.7 &67.3 &50.6 &57.8 		&.0513 &66.7 &58.7 &83.0 &68.8 		&.1117 &59.4 &59.4 &90.1 &71.6 		\\    
    Multilingual spec   &.1108 &54.4 &53.9 &83.5 &65.5 		&.0496 &60.8 &54.1 &75.5 &63.0 		&.1090 &63.1 &69.5 &62.6 &65.9 		\\
    Multilingual raw   	&.1080 &66.0 &66.7 &68.8 &67.7 		&.0539 &60.0 &53.3 &75.5 &62.5 		&.1141 &56.2 &62.1 &59.3 &60.7 		\\
    Fine-tuned raw   	&.1122 &64.1 &60.9 &85.5 &71.2 		&.0479 &65.8 &58.3 &79.2 &67.2 		&.1000 &73.1 &76.1 &76.9 &76.5 		\\
    \midrule
    \multicolumn{16}{c}{\textbf{Agreeableness}}																						\\
    \midrule
%     Single lang. spectrogram     &.095 & &64.7 &65.2 &65.0 &.038 &65.0 &65.0 &100 &78.8 &.064 &66.8 &71.4 &63.1 &67.0 \\
%     Single lang. raw audio       &.095 &62.8 &60.9 &77.2 &68.1 &.043 &64.5 &71.2 &66.7 &68.9 &.062 &72.0 &73.3 &74.8 &74.0 \\
%     Single lang. raw3    &.095 &62.8 &60.9 &77.2 &68.1 		&.043 &64.5 &71.2 &66.7 &68.9 			&.062 &72.0 &73.3 &74.8 &74.0 			\\
%     Single language raw   &.1009 &59.4 &59.5 &79.3 &68.0 			&.0643 &53.3 &54.3 &78.1 &64.1 			&.0743 &61.9 &63.2 &59.3 &61.1 			\\    
	Single lang. spec	&.0975 &59.6 &65.9 &53.6 &59.1		&.0518 &55.0 &61.4 &42.2 &50.0 		&.0679 &70.0 &74.6 &61.7 &67.6 		\\
    Single lang. raw   	&.1004 &58.8 &61.8 &63.7 &62.8 		&.0704 &46.7 &50.0 &6.2  &11.1 		&.0749 &60.0 &71.8 &34.6 &46.7 		\\    
    Multilingual spec   &.0989 &55.2 &56.1 &81.1 &66.3 		&.0586 &42.5 &45.6 &40.6 &43.0 		&.0744 &56.2 &55.7 &66.7 &60.7 		\\
    Multilingual raw   	&.0953 &60.9 &64.4 &63.2 &63.8 		&.0555 &46.7 &50.0 &68.8 &57.9 		&.0761 &51.9 &51.5 &84.0 &63.8 		\\
    Fine-tuned raw   	&.0969 &62.0 &60.8 &84.9 &70.9 		&.0508 &55.8 &55.9 &81.2 &66.2 		&.0661 &67.5 &72.3 &58.0 &64.4 		\\
	\midrule
	\multicolumn{16}{c}{\textbf{Conscientiousness}}																					\\
    \midrule
%     Single lang. spectrogram     &.116 & &63.8 &67.3 &65.5 &.054 &54.2 &- &0 &- &.059 &70.5 &73.0 &76.4 &74.7 \\
%     Single lang. raw audio       &.117 &64.1 &58.9 & 83.3 & 69.0 &.055 &77.3 &46.3 &69.1 &55.5 		&.055 &71.5 &72.7 &80.0 &76.2 \\
%     Single lang. raw3    &.117 &64.1 &58.9 & 83.3 & 69.0 	&.055 &77.3 &46.3 &69.1 &55.5 		&.055 &71.5 &72.7 &80.0 &76.2 		\\
%     Single lang raw   &.1225 &61.4 &59.8 &83.4 &69.7 		&.0610 &51.7 &43.9 &34.0 &38.3 		&.0759 &51.2 &93.3 &15.4 &26.4 		\\    
	Single lang. spec	&.1194 &61.2 &67.6 &52.1 &58.9		&.0509 &59.2 &62.5 &18.9 &29.0 		&.0629 &68.1 &78.6 &60.4 &68.3 		\\
    Single lang. raw   	&.1248 &59.4 &60.8 &66.4 &63.5 		&.0521 &55.8 &50.0 &37.7 &43.0 		&.0775 &60.0 &58.7 &100.0 &74.0		\\    
    Multilingual spec   &.1199 &56.3 &59.1 &57.6 &58.4 		&.0598 &49.2 &44.7 &64.2 &52.7 		&.0645 &61.9 &64.2 &74.7 &69.0 		\\
    Multilingual raw   	&.1160 &62.5 &65.8 &61.2 &63.4 		&.0564 &49.2 &44.6 &62.3 &52.0 		&.0646 &60.6 &61.9 &80.2 &69.9 		\\
    Fine-tuned raw   	&.1168 &62.6 &60.7 &84.2 &70.6 		&.0522 &50.0 &45.7 &69.8 &55.2 		&.0615 &65.0 &68.8 &70.3 &69.6 		\\
    \midrule
    \multicolumn{16}{c}{\textbf{Neuroticism}}																						\\
    \midrule
%     Single lang. spec &.107 & &66.9 &69.1 &68.0 			&.043 &35.0 &35.0 &100 &51.9 		&.073 &64.2 &64.6 &63.9 &64.2 		\\
%     Single lang. raw  &.108 &66.1 &65.6 &66.1 &65.9 		&.041 &22.1 &50.0 &59.5 &54.3 		&.066 &73.1 &72.3 &75.3 &73.7 		\\
%     Single lang. raw3 &.108 &66.1 &65.6 &66.1 &65.9 		&.041 &22.1 &50.0 &59.5 &54.3 		&.066 &73.1 &72.3 &75.3 &73.7 		\\
%     Single lang.raw   &.1119 &63.1 &62.6 &74.4 &68.0 		&.0461 &65.8 &65.0 &80.0 &71.7 		&.0765 &63.1 &63.4 &64.2 &63.8 		\\    
	Single lang. spec	&.1096 &64.3 &70.1 &56.4 &62.5		&.0427 &61.7 &62.3 &73.8 &67.6 		&.0722 &67.5 &68.8 &65.4 &67.1 		\\
    Single lang. raw   	&.1143 &61.1 &65.0 &56.8 &60.7 		&.0421 &68.3 &71.4 &69.2 &70.3 		&.0828 &55.6 &64.7 &27.2 &38.3 		\\    
    Multilingual spec   &.1121 &54.9 &55.6 &71.8 &62.7 		&.0484 &55.0 &63.4 &40.0 &49.1 		&.0789 &58.8 &59.0 &60.5 &59.8 		\\
    Multilingual raw   	&.1077 &64.8 &67.9 &62.8 &65.2 		&.0513 &51.7 &56.1 &49.2 &52.5 		&.0807 &60.0 &58.1 &75.3 &65.6 		\\
    Fine-tuned raw   	&.1102 &65.6 &62.6 &86.1 &72.5 		&.0426 &68.3 &74.5 &63.1 &68.3 		&.0731 &72.5 &70.3 &79.0 &74.4 		\\
    \midrule
    \multicolumn{16}{c}{\textbf{Openness to Experience}}																			\\
    \midrule
%     Single lang. spec &.104 &&65.6 &66.8 &66.2 			&.024 &64.2 &- &0 &- 				&.043 &59.1 &59.3 &69.3 &63.9 		\\
%     Single lang. raw  &.104 &65.4 &63.2 &69.6 &66.2 		&.027 &26.1 &43.2 &81.4 &56.5 		&.041 &62.2 &61.5 &74.3 &67.3 		\\
%     Single lang. raw3 &.104 &65.4 &63.2 &69.6 &66.2 		&.027 &26.1 &43.2 &81.4 &56.5 		&.041 &62.2 &61.5 &74.3 &67.3 		\\
%     Single lang. raw  &.1081 &62.7 &65.3 &59.5 &62.3 		&.0308 &62.5 &62.2 &73.0 &67.2 		&.0495 &55.0 &51.8 &93.5 &66.7 		\\    
	Single lang. spec	&.1048 &63.8 &67.7 &57.2 &62.0		&.0278 &57.5 &63.6 &44.4 &52.3 		&.0434 &60.0 &60.7 &48.1 &53.6 		\\
    Single lang. raw   	&.1099 &61.6 &66.9 &50.8 &57.7 		&.0353 &50.8 &52.0 &81.0 &63.4 		&.0502 &50.6 &49.2 &81.8 &61.5 		\\    
    Multilingual spec   &.1055 &54.1 &55.1 &60.3 &57.6 		&.0316 &51.7 &53.7 &57.1 &55.4 		&.0444 &52.5 &50.6 &55.8 &53.1 		\\
    Multilingual raw   	&.1024 &66.0 &67.5 &66.2 &66.9 		&.0283 &57.5 &57.9 &69.8 &63.3 		&.0433 &60.0 &57.5 &64.9 &61.0 		\\
    Fine-tuned raw   	&.1067 &64.1 &61.1 &84.3 &70.8 		&.0281 &56.7 &56.6 &74.6 &64.4 		&.0418 &66.2 &61.4 &80.5 &69.7 		\\
    \midrule
    \multicolumn{16}{c}{\textbf{Average over Traits}}																				\\
    \midrule
%     Single lang. spec &.106 & &65.5 &67.6 &66.6 					&.043 &53.0 &- &40.0 &- 				&.066 &65.9 &67.9 &69.8 &68.7 					\\
%     Single lang. raw  &.105 &64.9 &66.1 &68.9 &67.5 				&.042 &52.7 &55.6 &71.9 &61.9 			&.062 &70.6 &70.8 &77.6 &73.9 					\\
    Single lang. spec	&.1081 &62.5 &67.8 &55.4 &61.0 				&\textbf{.0436} &60.8 &63.6 &48.7 &53.0 &.0701 &67.0 &71.2 &61.6 &65.9 					\\
    Single lang. raw  	&.1127 &60.5 &64.4 &57.7 &60.5 				&.0502 &57.7 &56.4 &55.4 &51.3 			&.0794 &57.1 &60.8 &66.7 &58.4 					\\
    Multilingual spec   &.1094 &54.9 &56.0 &70.9 &62.1 				&.0496 &51.8 &52.3 &55.5 &52.6 			&.0742 &58.5 &59.8 &64.1 &61.7 					\\
    Multilingual raw  	&\textbf{.1059} &64.0 &66.5 &64.5 &65.4 	&.0491 &53.0 &52.4 &65.1 &57.6 			&.0758 &57.8 &58.2 &72.8 &64.2 					\\
    Fine-tuned raw   	&.1086 &63.7 &61.2 &85.0 &\textbf{71.2} 	&.0443 &59.3 &58.2 &73.6 &\textbf{64.3} &\textbf{.0685} &68.9 &69.8 &73.0 &\textbf{70.9}\\
    
  \end{tabular}
  }
\end{table*}

\subsection{Experimental setup}
To build the test sets, for the corpora which included different speakers of different genders. For the German, Italian and Serbian corpora one speaker of each gender was used as the test set. For the other three corpora we could not apply speaker separation. In the Spanish and Estonian corpora contained too few speakers for each gender: one male and female the former, and only one female speaker the latter. In the English dataset instead most of the samples did not include any information about the speaker identity. In any case the overall number of speakers and samples in this language was much greater than the other language corpora, since it includes data from a large number of sources. For these three datasets around $10\%$ of samples of each emotion class were taken as test set. The detailed division among training and test set is reported in Table \ref{emotion_speakers}.  The test set was kept the same during the multiclass and fine-tuning training phases, as well as with various network configurations. In order to tune the network structure and hyperparameters, and determine the early stopping condition, a subset of the training set of $10\%$ was each time randomly taken as the development set.

Each audio sample was transformed into \verb|wav| format at 16 bits and downsampled to $8\,\text{kHz}$ with \verb|sox|\footnote{http://sox.sourceforge.net}. To keep the input range of every sample small and avoid parameter overflowing during training, a constant value of $k = 5 \cdot 10^{-4}$ was multiplied to every input audio sample. The $k$ value was chosen in order to approximately normalize the overall standard deviation to 1. The volume randomization hyperparameter $a$ (see Eq. \ref{volnorm}) was set to 1.5.

We apply four convolutional layers after the first feature extraction layer, the first layer with a kernel size of 8 and a stride of 2, and each subsequent ones with a kernel size of 4 and a stride of 2. This means that each layer from the first to the last analyzes increasingly larger time spans starting from $25\,\text{ms}$. To train our CNN we applied standard backpropagation with Adam optimizer~\cite{kingma2014adam}. The initial learning rate was set to $10^{-4}$, and halved once after the first 25 epochs and subsequently after another 15 epochs. We stopped training when the error on the development set began to increase. During the global multiclass training a minibatch size of 2 was used, while in the single class and fine-tuning we used a minibatch size of 1.

\subsection{Results}
Results of our experiments on multilingual emotion detection are shown in Table \ref{res_emotion}. They are represented in terms of precision, recall and F-score over each emotion and language. We obtained an average F-score of 67.7\% (68.5\% after fine-tuning the last layer) across all the languages using our raw waveform CNN trained on multiple languages. We obtained an average of 55.2\% with the same model trained on a single language, 58.2\% from the multilingual spectrogram baseline and 60\% from the same baseline trained on single languages. Overall, this yields a relative improvement of 12.8\% of the multilingual raw waveform CNN over the second best model, the spectrogram CNN trained on individual languages.  

\section{Personality Recognition Experiments}
\label{personality_experiments}

\subsection{Corpora}
For the personality recognition task we use three different languages datasets: a bigger one in English and two smaller ones in Mandarin and French. Each sample from each dataset is annotated with five continuous scores between 0 and 1 (for the Big Five personality traits). Each dataset is recorded at a sampling rate of at least $8\,\text{kHz}$. The datasets are:

\begin{itemize}
	\item \textbf{English - ChaLearn Looking at People 2016 Apparent Personality Analysis (APA) Dataset}~\cite{ponce2016chalearn}: consists of 8,000 clips of around 15 seconds, taken from YouTube blogs with diverse conversational content. The videos are labeled by Amazon Mechanical Turk workers. Audio clips are extracted from the videos.
% 	\item \textbf{English - BIT Speaker Personality Corpus (BSPC)}~\cite{zhang2017social}: consists of 347 male and 186 female clips taken from 120 English talk shows, extracted at 16 bits with a sampling rate of 44.1 kHz. Clip length varies from 9 to 13 seconds. The utterances are labeled by student workers by filling in a NEO-PI-R questionnaire for the speakers, with three judges per audio clip. Male and female clips were combined for the experiments in this work, and downsampled to 8 kHz.
	\item \textbf{Mandarin Chinese - Beijing Social Personality Corpus (BSPC)}~\cite{zhang2017social}: consists of 258 male and 240 female clips taken from 70 Chinese talk shows. Clip length varies from 9 to 13 seconds. The utterances are labeled by three student workers each by filling in a standard NEO-PI-R personality inventory for the speakers.
	\item \textbf{French - SSPNet Speaker Personality Challenge}~\cite{mohammadi2012automatic}: consists of 640 clips taken from French radio shows. Each clip is labeled by 11 unique judges. Final scores are taken as the average of the scores of these judges.
\end{itemize}

% \begin{table}[th]
%   \label{tab:personality_corpora}
%   \centering
%   \caption{Overview of training labels of datasets used for personality recognition. Datasets have different label distributions, which will need to be normalized.}
%   \scalebox{0.85}{
%   \begin{tabular}{ r ccccc }
%      				&Extr. &Agre. &Cons. &Neur. &Open. 	\\
%     \midrule
%     \multicolumn{6}{c}{\textbf{ChaLearn APA - English}} \\
%     \midrule
%     	$Avg.$		&0.476 &0.549 &0.523 &0.521 &0.567	\\
%     	$Std.$ 		&0.152 &0.136 &0.155 &0.153 &0.147 	\\
%    \midrule
%    \multicolumn{6}{c}{\textbf{BSPC - Chinese}} 			\\
%    \midrule
%     	$Avg.$    	&0.545 &0.584 &0.653 &0.414 &0.519	\\
%     	$Std.$      &0.063 &0.071 &0.066 &0.058 &0.036	\\
%    \midrule
%    \multicolumn{6}{c}{\textbf{SSPNet - French}} 		\\
%    \midrule
%     	$Avg.$    	&0.585 &0.538 &0.634 &0.433 &0.529 	\\
%     	$Std.$      &0.138 &0.093 &0.083 &0.101 &0.055 	\\        
% %    \midrule
% %    \multicolumn{6}{c}{\textbf{myPersonality - English}} \\
% %     		$Avg.$    	&0.503 &0.711 &0.587 &0.524 &0.727	\\
% %     		$Std.$      &0.231 &0.179 &0.183 &0.215 &0.156	\\        
% %     		$MAE$      	&0.148 &0.121 &0.121 &0.170 &0.120	\\        
%   \end{tabular}
%   }
% \end{table} 

It's important to note that the distributions (means and standard deviations) of trait scores differ per dataset. Especially the spread in scores for the Chinese dataset is very small. To combine all data for training, the labels thus need to be normalized.

%Table~\ref{tab:personality_corpora} shows the mean and standard deviation of the training set labels of each corpus. 
%The averages will later be used as decision border for the 2-class evaluation. 
For the English dataset we use the pre-defined ChaLearn Validation Set (2,000 samples) as test set. For Mandarin, we take 60 samples each from male and female speakers, which results in 120 samples in total. For French, we take out 160 random samples to serve as test set. As the development set we used 10\% samples from each corpus.

\subsection{Experimental setup}
For the personality recognition experiments we used \textbf{four} convolutional layers in the CNN. Everything else is identical to the architecture used for the emotion recognition experiments. We pre-processed the input samples and trained the network mostly in the same way, and with the same single and multilingual experiments, as described for emotion in the previous section. However, an important exception is represented by the labels. Due to the difference in label distributions (mean and spread), across the three datasets we rescaled all training labels to have the same mean and standard deviation before training. We assumed the labels distribution for each personality trait as a Gaussian random variable. At evaluation time, the output predictions were inversely converted back to the original distribution for each individual language. We trained the model with a regression cost function by minimizing the Mean Squared Error between model prediction and ground truth:
\begin{equation}
	\text{MSE} = \frac{1}{N} \sum^N_{i=1} (p_i-g_i)^2
\end{equation}
where $p_i$ is the vector of the five trait predictions for a given sample $i$ and $g_i$ is the vector of the five ground truth trait values for that sample. Another difference is the higher learning rate of $2 \cdot 10^{-4}$.

We evaluate the model both from a regression point of view, evaluating the Mean Absolute Error (MAE) between the prediction and the ground truth for each trait, and from a classification point of view by turning the predictions and corresponding labels into binary classes using the average of each trait as the boundary between the two classes. In this setting we compute classification accuracy, precision, recall and F-score.

\begin{figure*}[ht!]
\centering
  \subfloat{%
    \includegraphics[width=0.5\columnwidth,trim={0 1.5cm 0 3cm},clip]{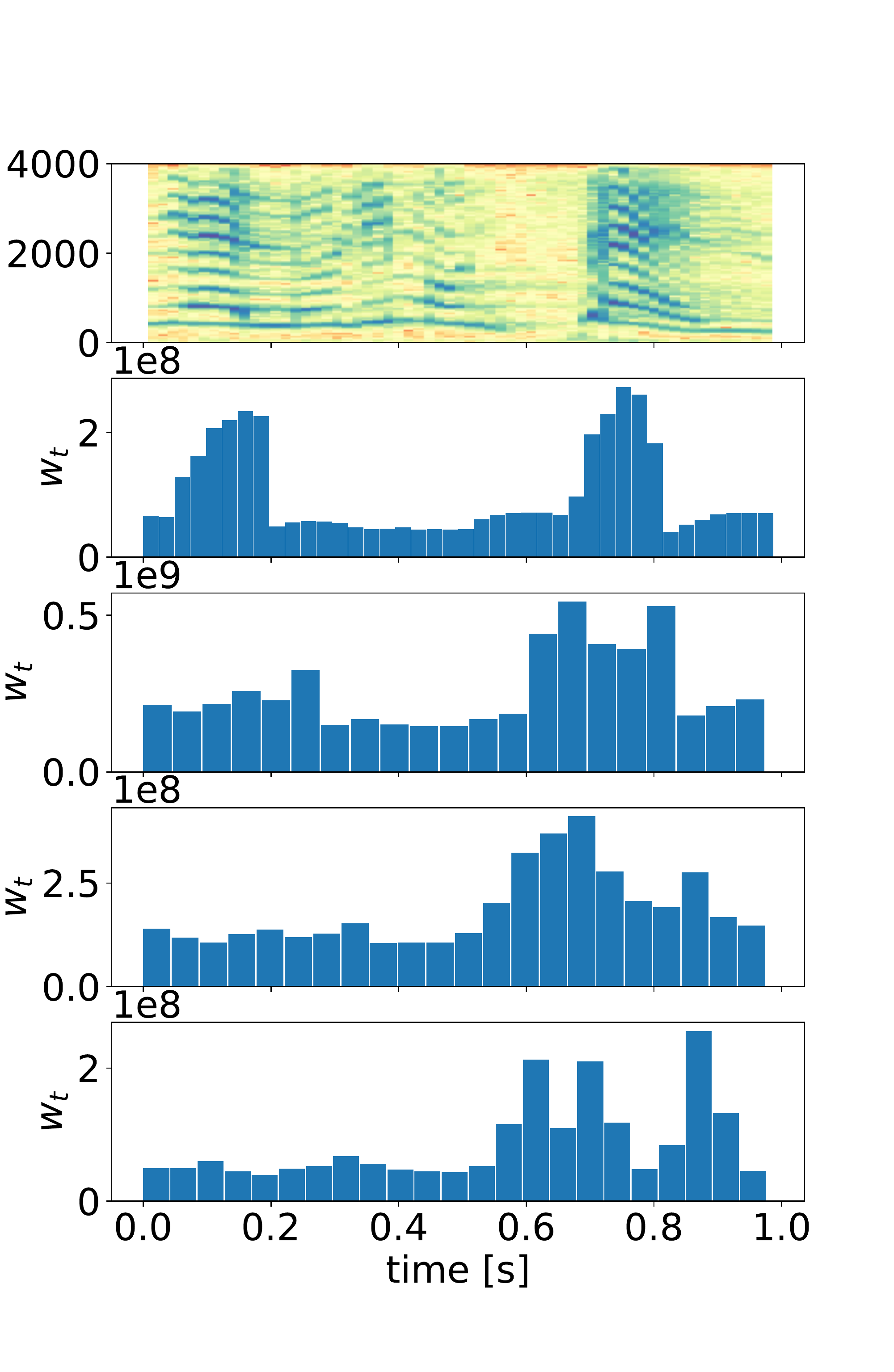}%
  }
  \subfloat{%
    \includegraphics[width=0.5\columnwidth,trim={0 1.5cm 0 3cm},clip]{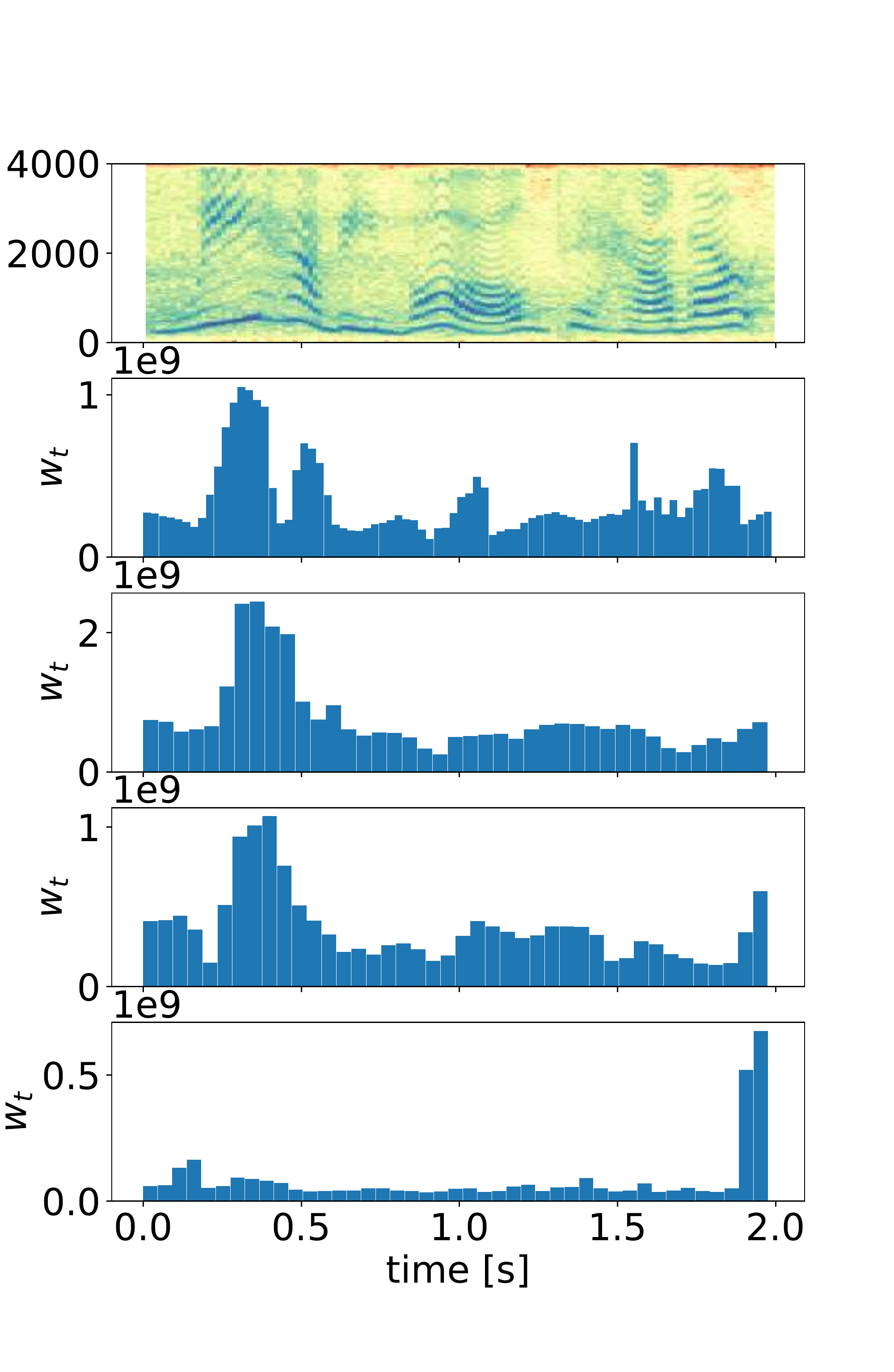}%
  }
  \subfloat{%
    \includegraphics[width=0.5\columnwidth,trim={0 1.5cm 0 3cm},clip]{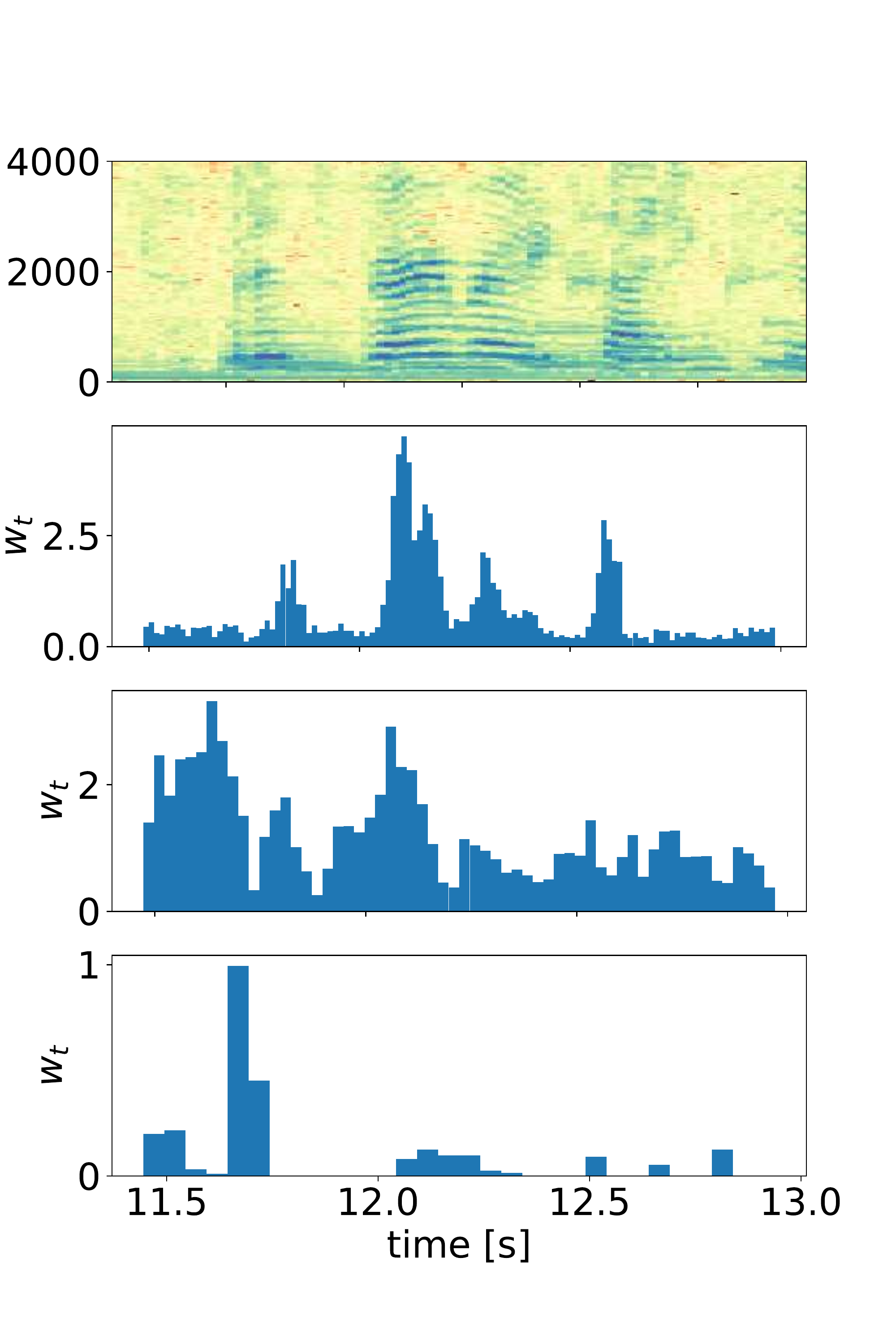}%
  }
  \subfloat{%
    \includegraphics[width=0.5\columnwidth,trim={0 1.5cm 0 3cm},clip]{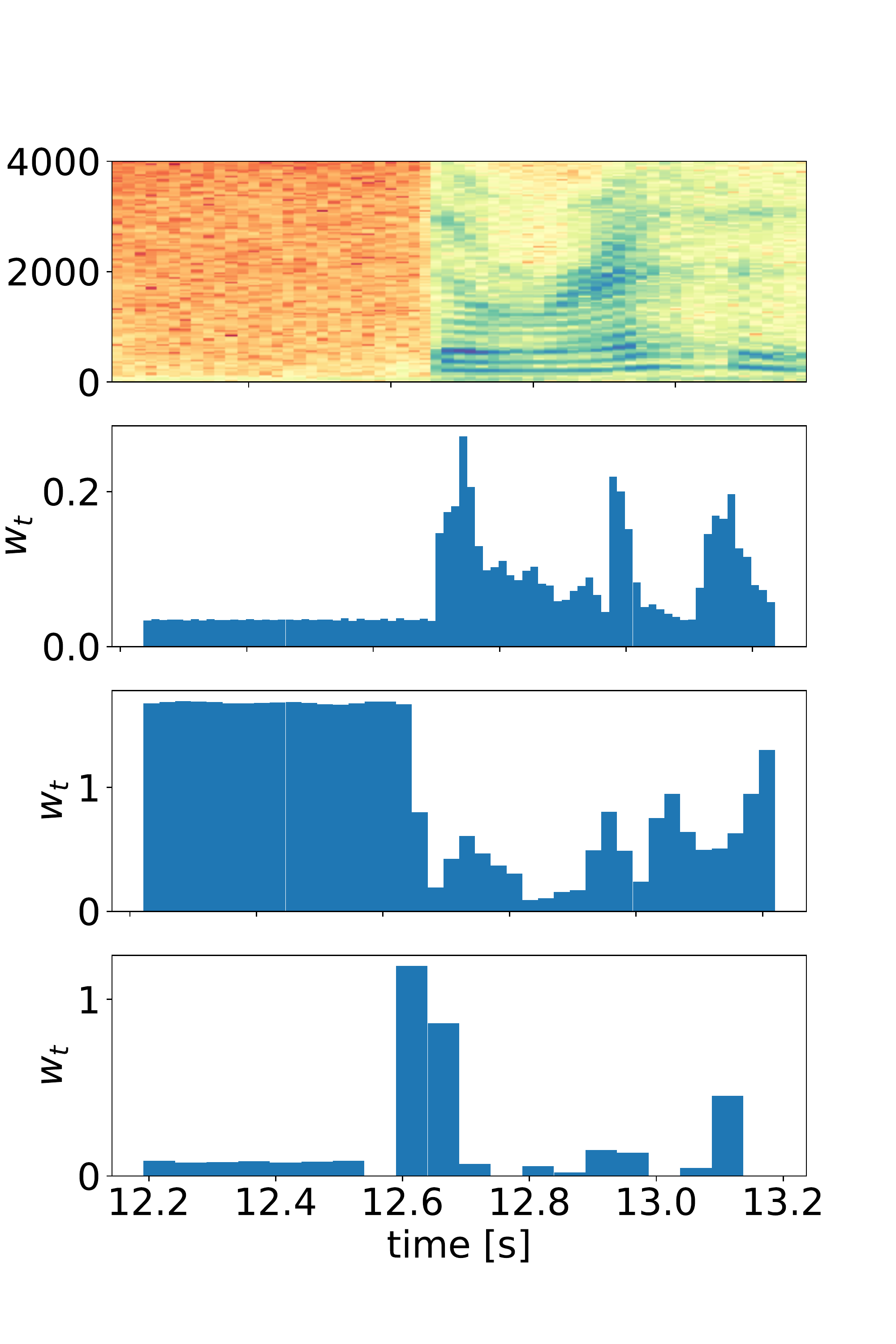}%
  }
\caption{Spectrogram representation of short speech samples from the corpus (top), and relative RMS activation over time of the intermediate layers  for the emotion (left two) and personality (right two) networks. Figures show how the higher network layers activate on the voiced parts with different patterns, especially when there is a change in prosody. For emotion, silences are mostly ignored, whereas for personality one layer also activates heavily on longer durations of silence.}
\label{activation}
\end{figure*}

\begin{figure}[t!]
  \centering
  \includegraphics[width=\columnwidth, trim={0 0.6cm 0 0.6cm}, clip]{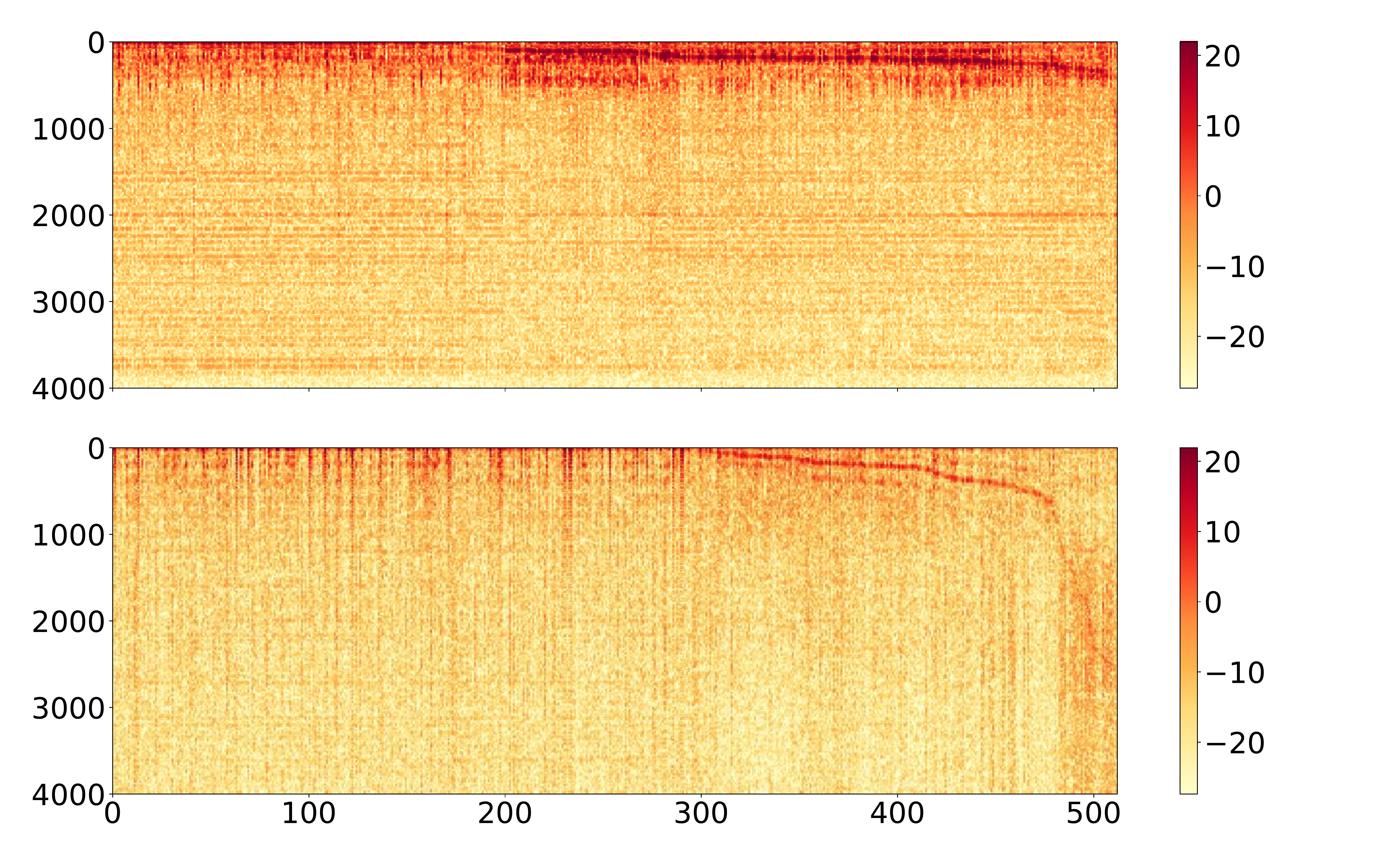}
  \caption{Frequency response of each of the 512 filters (horizontal axis) of the first-layer of the CNN for emotion recognition. Above are shown the filters applied to the raw signal, below those applied to the squared signal. It is evident how energy (left) and pitch (middle) are the main features extracted for emotion recognition by the CNN.}
  \label{filters_emotion}
\end{figure}

\begin{figure}[t!]
  \centering
  \includegraphics[width=\columnwidth, trim={0 0.6cm 0 0.6cm}, clip]{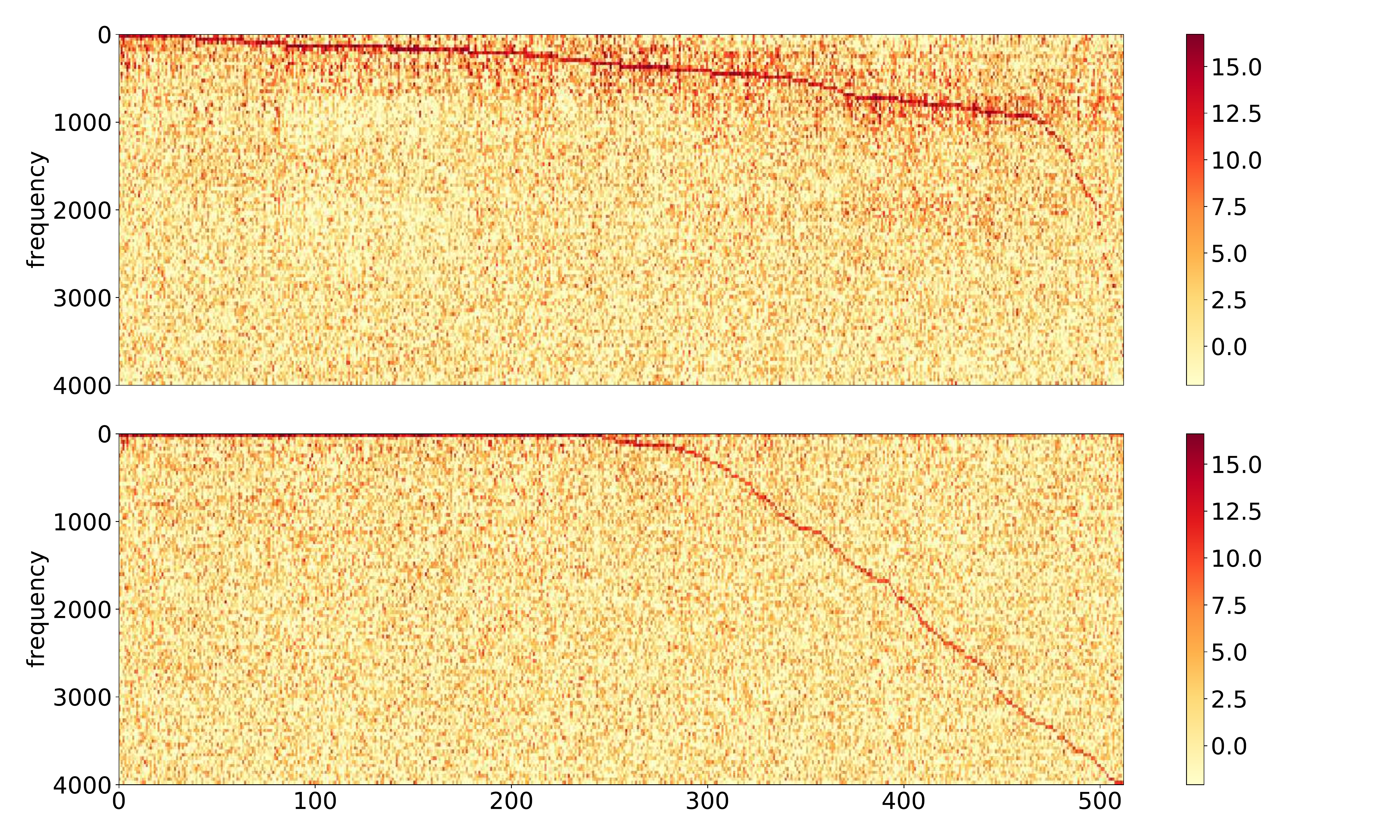}
  \caption{Frequency response of each of the 512 filters (horizontal axis) of the first-layer of the CNN for personality recognition. Above are shown the filters applied to the raw signal, below those applied to the squared signal. In this case the network extracts a wider feature set than in the emotion detection case. These features include energy, pitch, contour variations and also frequency components between 500 and 1000 Hz.}
  \label{filters_personality}
\end{figure}

\begin{figure*}[ht!]
\centering
  \subfloat{%
    \includegraphics[width=\columnwidth, trim={3.5cm 3.5cm 3.5cm 3.5cm}, clip]{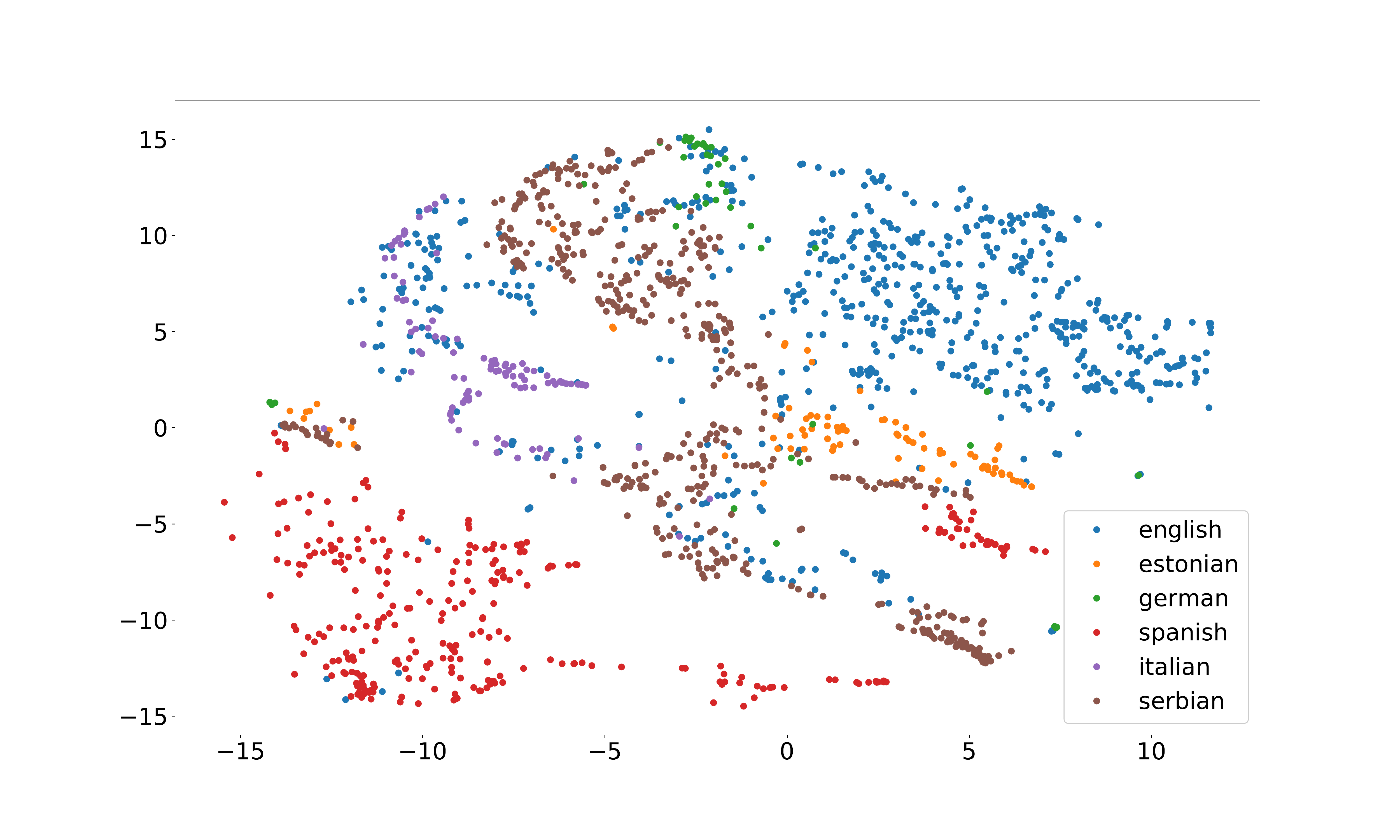}%
  }
%   \subfloat{%
%     \includegraphics[width=\columnwidth, trim={3.5cm 3.5cm 3.5cm 3.5cm}, clip]{./tsne_emotion_language_second.pdf}%
%   }\\
  \subfloat{%
    \includegraphics[width=\columnwidth, trim={3.5cm 3.5cm 3.5cm 3.5cm}, clip]{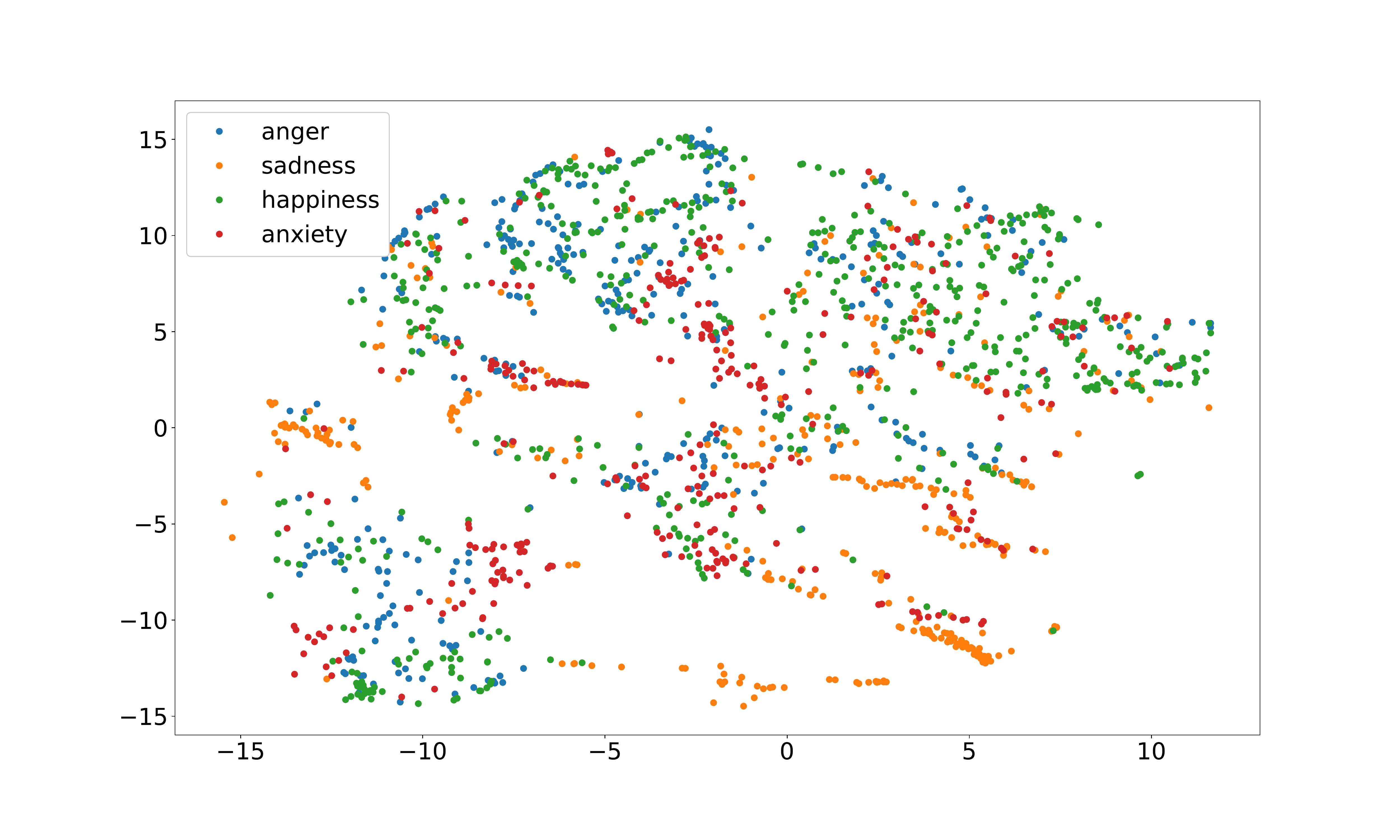}%
  }\\
\subfloat{%
    \includegraphics[width=\columnwidth, trim={3.5cm 3.5cm 3.5cm 3.5cm}, clip]{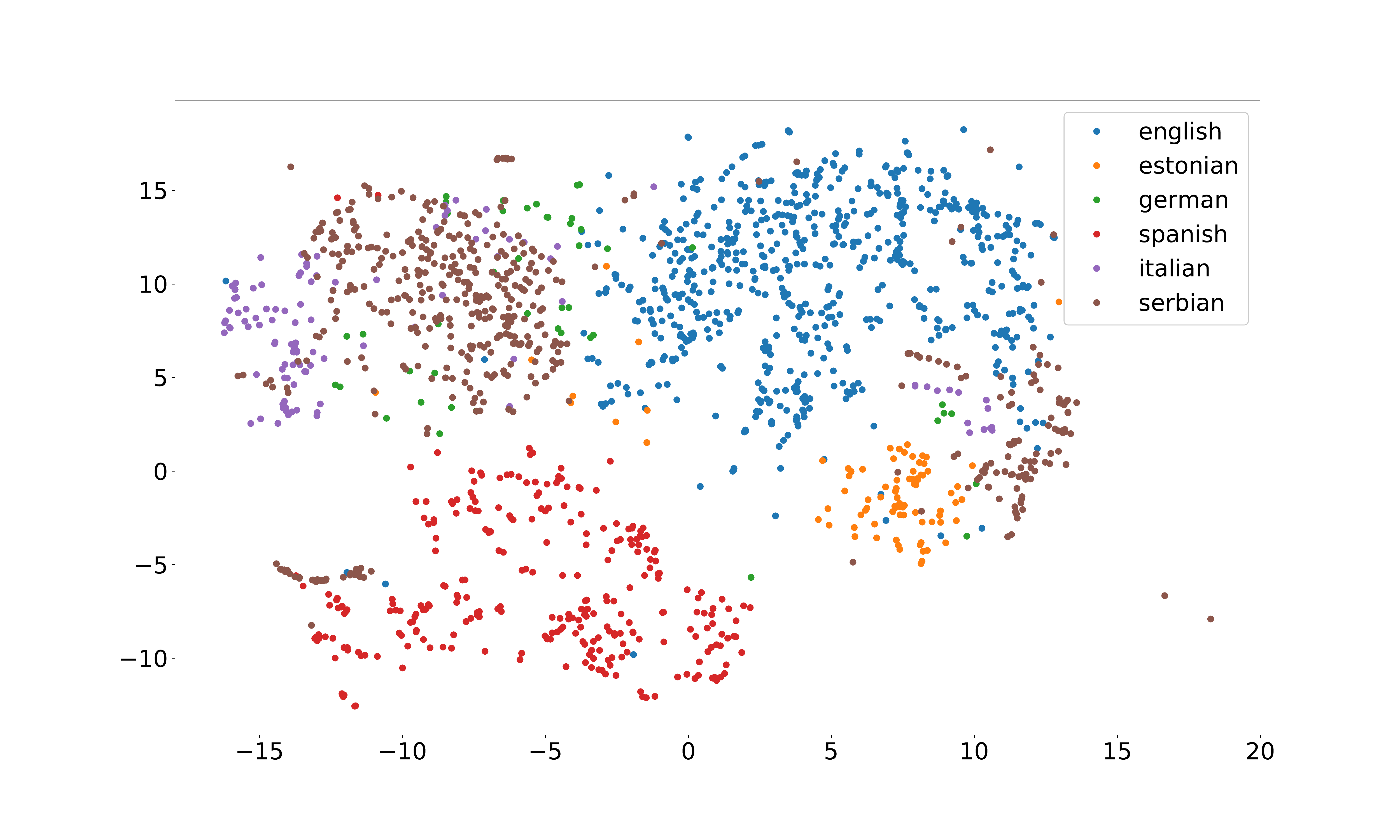}%
  }
%   \subfloat{%
%     \includegraphics[width=\columnwidth, trim={3.5cm 3.5cm 3.5cm 3.5cm}, clip]{./tsne_emotion_language_second.pdf}%
%   }\\
  \subfloat{%
    \includegraphics[width=\columnwidth, trim={3.5cm 3.5cm 3.5cm 3.5cm}, clip]{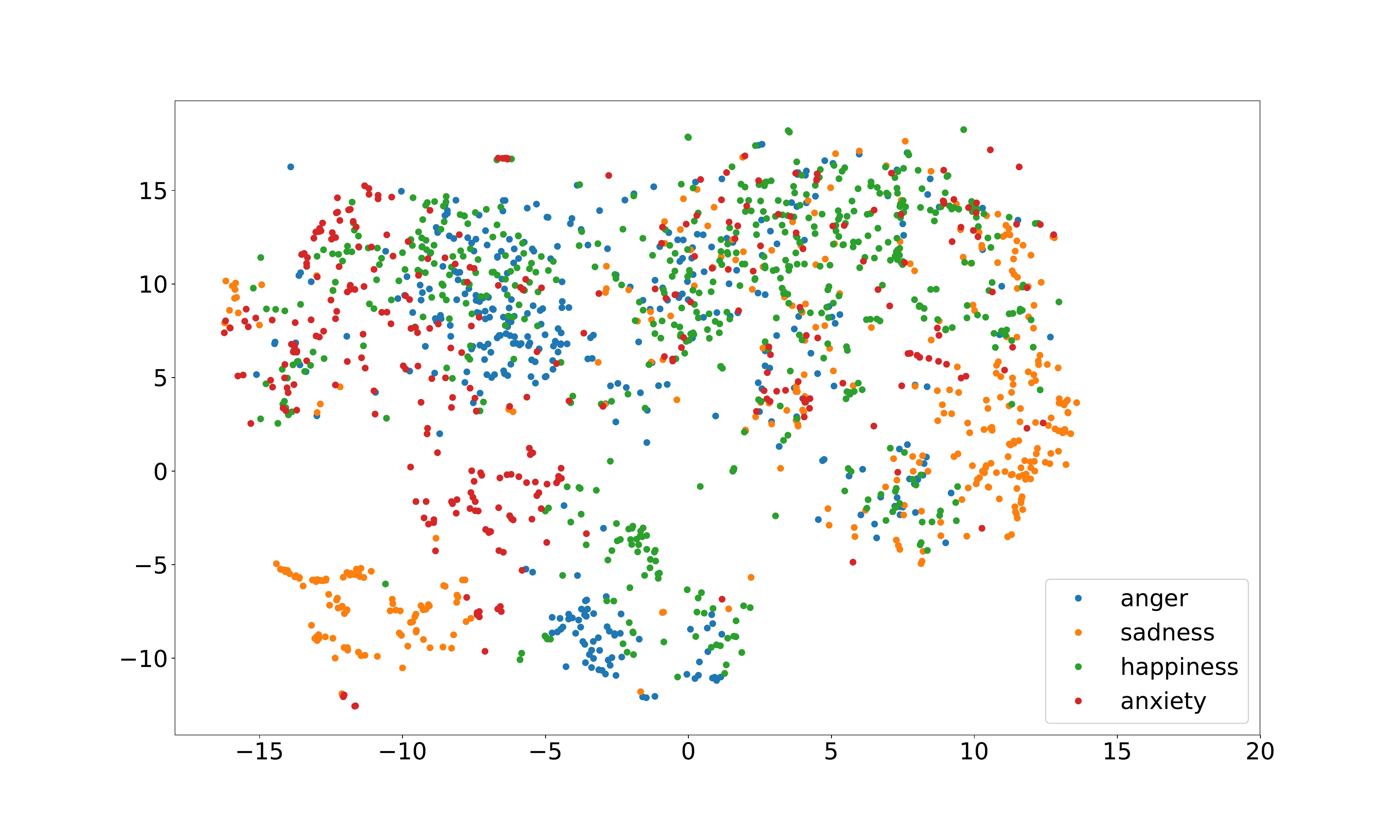}%
  }\\
  \subfloat{%
    \includegraphics[width=\columnwidth, trim={3.5cm 3.5cm 3.5cm 3.5cm}, clip]{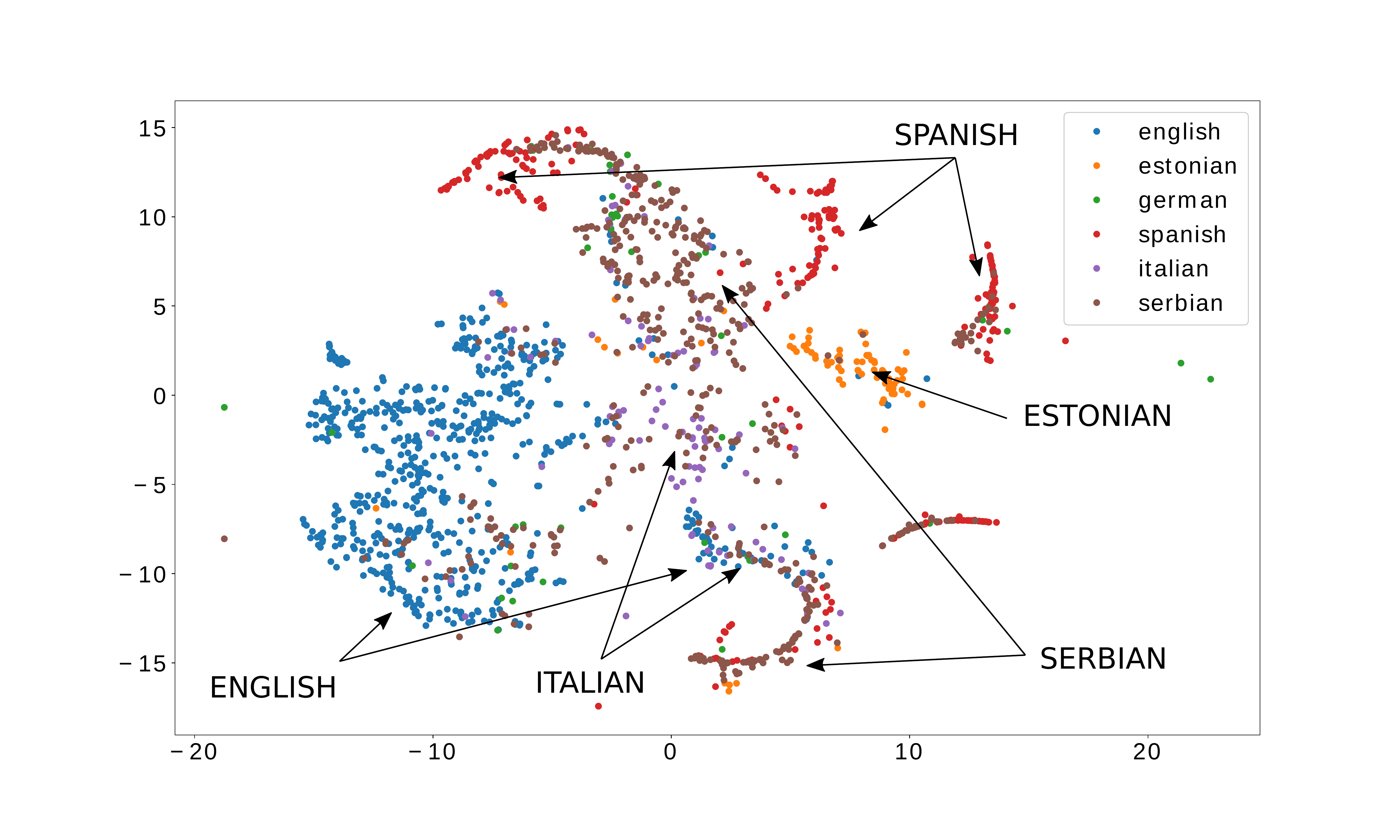}%
  }
  \subfloat{%
    \includegraphics[width=\columnwidth, trim={3.5cm 3.5cm 3.5cm 3.5cm}, clip]{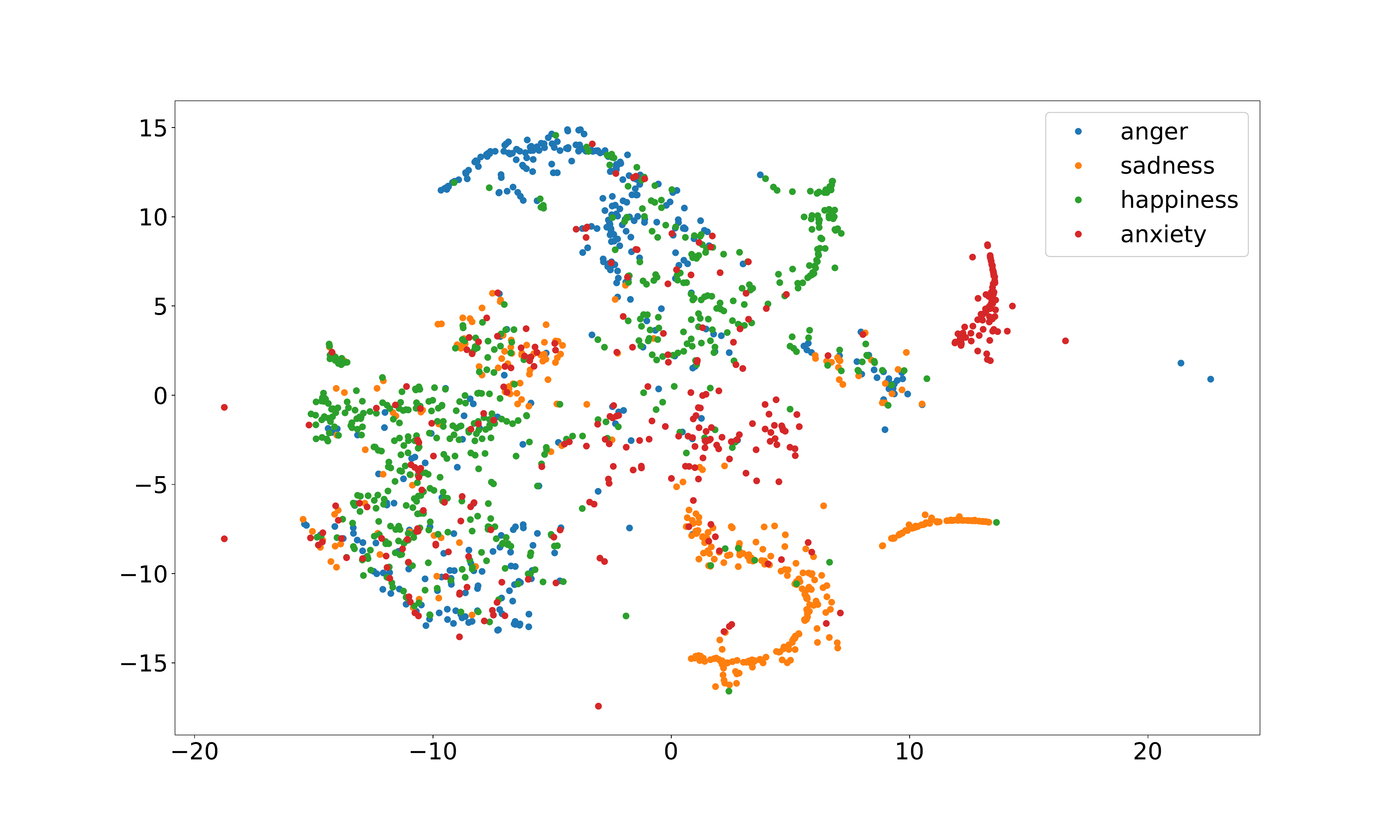}%
  }
\caption{t-SNE projection of outputs from average pooling after the first and fifth convolutional layers, and the last fully connected layer of the emotion recognition CNN. The left column shows the sample points with the languages highlighted in different colors. The right column shows the same projection with emotion labels highlighted in different colors. Higher convolutional layers tend to cluster samples according to the language compared to the first layer. The last fully connected layer shows stronger emotion clustering instead. Some languages, in particular Spanish, Serbian, German and some English samples, seem to be clustered together according to the emotion, thus interacting with each other to build the final predictions.}
\label{tsne_emotion}
\end{figure*}

\begin{figure*}[ht!]
\centering
  \subfloat{%
    \includegraphics[width=\columnwidth, trim={0 0.8cm 0 1cm}, clip]{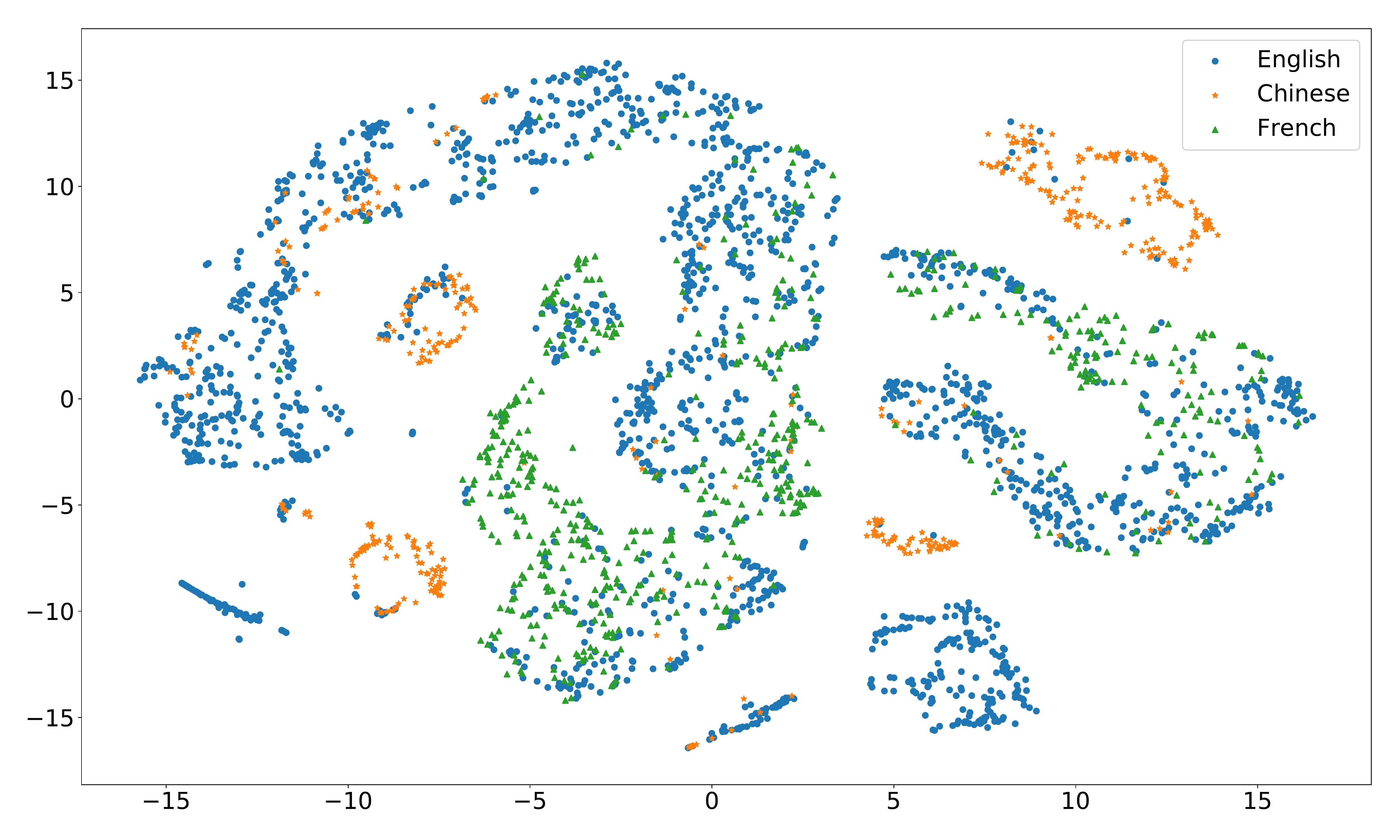}%
  }
  \subfloat{%
    \includegraphics[width=\columnwidth, trim={0 0.8cm 0 1cm}, clip]{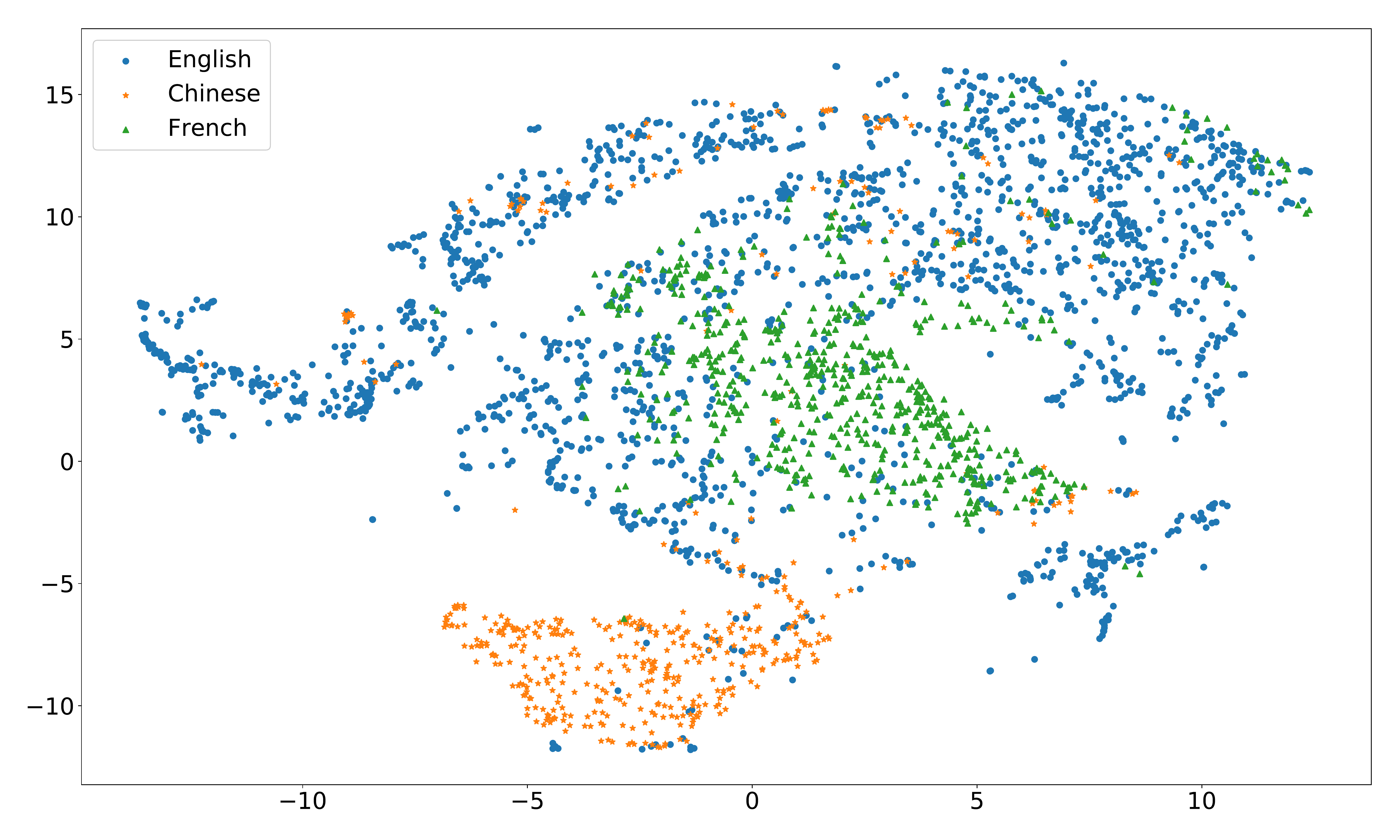}%
  }\\
  \subfloat{%
    \includegraphics[width=\columnwidth, trim={0 0.8cm 0 1cm}, clip]{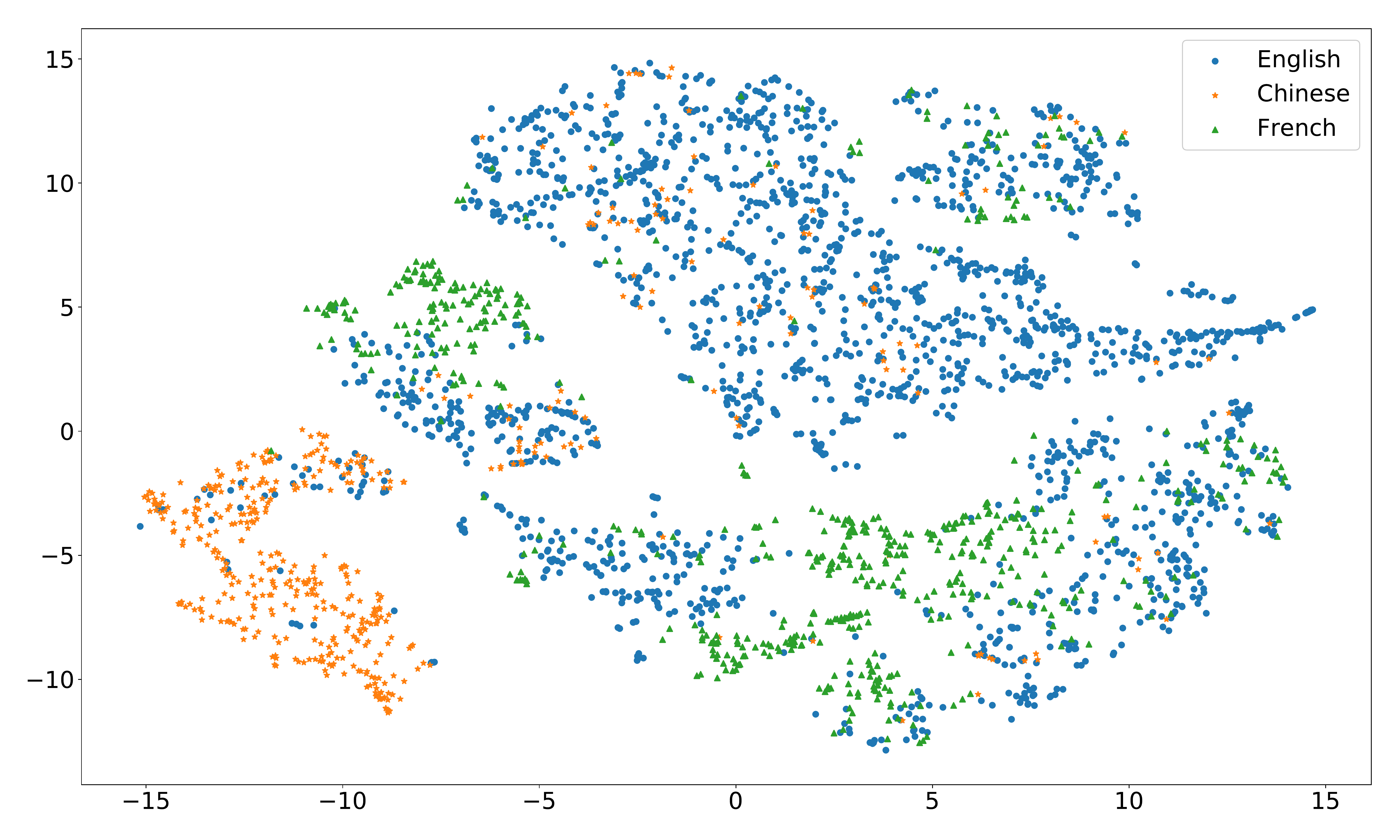}%
  }
  \subfloat{%
    \includegraphics[width=\columnwidth, trim={0 0.8cm 0 1cm}, clip]{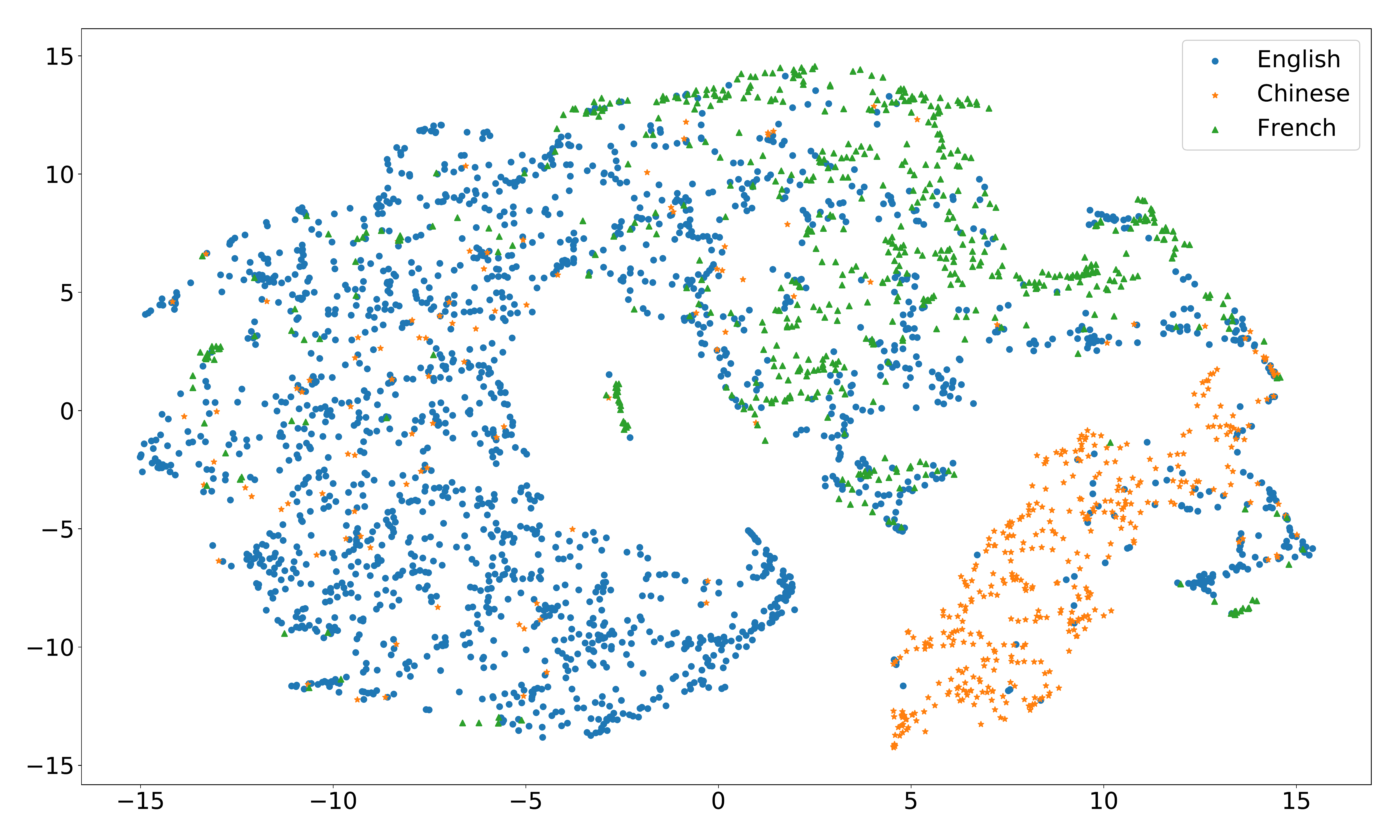}%
  }
\caption{t-SNE projections of average pooling after each CNN layer for personality recognition, from first to fourth (left to right, top to bottom). The language of each sample is highlighted in different colors and symbols. Language clusters appear especially after the second layer, and get more distinct after each layer. Chinese samples tend to have more distinct clusters than the other two languages (English and French).}
\label{tsne_personality}
\end{figure*} 

\subsection{Results} 
Results on each corpus, including the average over each trait for each language, are shown in Table~\ref{tab:res_personality}. The fine-tuned multilingual model performs best on the test sets in terms of F-score. For the multilingual model using raw waveforms, we obtained an average F-score over the three languages of 62.4\%. Training this same model on each language individually resulted in an average F-score over the languages of 56.7\%. Using a spectrogram instead of raw waveforms gives an average F-score of 58.8\%. Thus, our multilingual efforts show a relative improvement of 6.3\% over the spectrogram approach and 10.1\% over the single language approach. 

\section{Discussion}
\label{discussion}

\subsection{Affect recognition performance}
Results obtained for both emotion and personality recognition show that in all cases the multilingual training with raw waveforms input outperforms both the spectrogram input and the single language training. In some cases, like the German or Serbian emotion corpora, and the Chinese personality dataset, the improvement is particularly significant. Another evident result, in particular on the emotion experiments, is that using raw waveforms improves the performance of the multilingual training, while on the other hand the spectrogram input is better on the single language case. Fine-tuning of the last layers helps in most cases achieving an improvement, although in a minority of cases it is not that beneficial. It seems less useful when the datasets are larger than average (the two English datasets) or very small (the emotion German corpus).

Regarding the emotion recognition task, there is no particular emotion that is easier or more difficult across all the languages. Some emotions in specific languages are sometimes mistaken, for example in the English dataset \emph{anxiety} is often classified as \emph{sadness}, or German \emph{happiness} as \emph{anger}. These misclassifications are often related to the specific corpus characteristics.

\subsection{Low-level feature selection layer}
The first layer of our CNN has the role of extracting low-level features from the raw waveforms. It is important to visualize and understand which kind of features are extracted, how much these features correspond with those used in traditional feature-based approaches~\cite{schuller2009interspeech}, and whether something new or unusual is extracted.

To visualize the first layer we follow a similar procedure as used in~\cite{golik2015convolutional, bertero2017icassp}. We consider each row of the parameter matrix $\mathbf{W}^c$, which represents a filtering function applied to each convolution window and whose output is then summed together before the application of the non-linearity. We transform each filter to the frequency domain, taking the absolute values of the FFT:
\begin{equation}
  F(\mathbf{W}^{(1)}_i) = |\text{FFT}(\mathbf{W}^{(1)}_i)|
\end{equation}
where $i$ is the filter index. Each FFT coefficient represent the activation of the filter to each frequency. We do this for both the raw waveform channel and the squared signal channel. The activation values have been converted to logarithmic scale with the following function:
\begin{equation}
  a(i,f) = 20\,\log_{10}(F(W^{(1)}_{i,f}))
\end{equation}
To better show the filter contributions we sorted them according to the frequency with the highest activation, in ascending order. Figures~\ref{filters_emotion} and~\ref{filters_personality} show the filter activation pattern respectively for the emotion recognition and personality recognition experiments.

In the emotion recognition experiments, three kinds of features are evident from the plots. Approximately one-third of the filters applied to the raw waveform, and more than half of those applied to the squared value have their peak at $0\,\text{Hz}$. This first set of filters is likely capturing the signal energy. A second set of filters in reverse proportion over raw signal and squared signal channels has instead its peak over a narrow range of low frequency values, between $0$ and $250\,\text{Hz}$. Those filters act as pitch detectors, and this is compatible with the fact that the average human pitch frequency lies below $250\,\text{Hz}$ for both males and females~\cite{baken2000clinical}. 

It is interesting to note what happens for frequencies above $250\,\text{Hz}$. In the original waveform signal input channel, very few filters have their central frequency between 250 Hz and 500 Hz, and the higher frequencies in the spectrum are almost ignored. This may be due to an amount of emotion data available too small to capture effectively further information at higher frequency, or might suggest the hypothesis that high frequency components do not carry useful information for emotion detection. If the latter hypothesis is confirmed, there would be no need to use wide-band audio to improve the performance on emotion detection. However, in the squared signal input channel, a small number of filters extend above 500 Hz. These filters may capture the local amplitude variations of the signal, particularly frequent in angry speech. They may also learn an amplitude normalization function to apply to the signal, to remove the effect of variable amplitude levels at input (often due to non-uniform recording volume across samples). This hypothesis is supported by the observation that most filter activate dominantly on $0\,\text{Hz}$.

% The filter intensity in this channel is usually lower than on the raw signal channel (as shown in Figure \ref{filters_average_emotion}). 

For personality recognition, a similar observation can be made about energy (activations at $0\,\text{Hz}$) and pitch (activations between $0$ and 250 Hz). On the other hand, a third of the filters activate between 500 and 1000 Hz, higher than the cutoff frequency for emotion. These higher activation frequencies also result in about a third of the filters for the squared signal input channel activating strongest at higher frequencies. Since the squared signal is likely used for internal normalization, this may indicate a more complex normalization for higher frequency components in the signal.

\subsection{Intermediate convolutional layers}
As mentioned in section~\ref{methodology}, the second to last convolutional layers are aimed at combining features at supra-segmental level and, among others, selecting the most salient phonetic units that may carry the emotion or the personality information. 

To visualize the contribution of these layers over a few examples, we estimate from the average pooling vectors a weighing factor $w_t$ to each time window taking the RMS value of the difference between the average pooling values and the element-wise average, in the following way:
\begin{equation}
	w_t = \sqrt{\dfrac{\sum_i\left(x^t_i - \overline{x}_i \right)^2}{N}}
\end{equation}
where $i$ is the vector element index, N the vector length (512 in the emotion detection experiments) and $\overline{x}_i$ the element-wise average among all time instants. A high $w_t$ means that some of the filters have a different value than the average for that specific time-frame, and are more likely to contribute to the final classification.

Figure \ref{activation} shows the activation of the intermediate convolutional layers over speech segments randomly taken from the corpus respectively for emotion and personality. For emotion, the uniform low activation pattern over the silence regions shows these do not usually carry any emotion-related information. For personality, filters do activate over silence, indicating these regions are correlated with personality. The intermediate layers activate strongly over voiced regions, especially when there is a prosodic change, such as energy or pitch variations. The activation pattern is often similar among the layers, but it is slightly more sparse toward the upper layers. This signals that upper layers tend to select the most important features extracted by the lower layers.

The behavior of each layer after the average pooling operation, and the final fully-connected layer, is also worth noticing. We projected the output of each intermediate layer into a two-dimensional space through a t-SNE transformation~\cite{maaten2008visualizing}. The output for the intermediate layers of the emotion and personality detection networks are respectively shown in Figures~\ref{tsne_emotion} and~\ref{tsne_personality}, highlighting the language of each sample. The figures illustrate that later convolutional layers are grouping each source language into its own cluster, with more defined cluster boundaries going upwards in the layers hierarchy. It seems that, through suprasegmental feature analysis, the network is automatically learning a specific affect model for each single input language. In the emotion recognition experiments (Fig.~\ref{tsne_emotion}, first and second rows) this pattern is very clearly shown by the t-SNE for all languages, except German due to the low amount of samples in that language. This pattern is also shown in the personality recognition experiments (Fig.~\ref{tsne_personality}. We also note that the Mandarin Chinese cluster is clearly distinct from the English and French clusters, which can be explained by the different cultural factors between Europe and China which may affect personality and its perception by annotators. Another factor contributing to this might be that English and French are much more similar phonetically than they are to Mandarin Chinese. In the emotion recognition case instead, Spanish seems to have a more distinct cluster. This dataset also yields the best average performance, which could be because it is acted and emotions are very clearly expressed. 

Overall, these figures show that, as we expected from previous multilingual acoustic modeling~\cite{fung2008multilingual,li2011asymmetric}, languages do share common features in the low, signal processing level, while they tend to have distinct characteristics at the higher, perhaps phonetic level. All these components are sent to the final classification layers, allowing the network to use both the common and different characteristics of the languages and use them to improve the final predictions. This is evident in the t-SNE representation of the emotion recognition last layer (Fig. \ref{tsne_emotion}, last row). Some groups of languages, in particular Serbian, Spanish, German and some English and Italian samples share emotion clusters. This could indicate that emotions from different languages have similar representations inside the network, thus explaining why adding data from other languages improves the model's performance. The exception to this is Estonian, which has a very different root from the other languages. We do not show these projections for personality recognition, as the regression nature of this task prevents clear clusters to form.

\section{Conclusion}
\label{conclusion}
In this paper, we proposed a universal end-to-end affect recognition model using convolutional neural networks. It is able to automatically extract features from narrow-band raw waveforms and detects emotions and personality traits regardless of the input language, whose characteristics are automatically learned and distinguished. We have obtained significant improvements both over a spectrogram baseline (+12.8\% for emotion and +6.3\% for personality), and training it with a multilingual setting as opposed as each single input language (+12.8\% for emotion and +10.1\% for personality). That is, we have shown that using raw waveforms yields higher performance than using spectrograms as input, and that training on multiple languages increases evaluation performance on each individual one in comparison to training separate models for each language. We have furthermore shown how the first convolutional layer in the model extracts low level features from the audio sample, while higher layers activate on prosodic changes and learn language-specific representations. 

We have shown that universal affect recognition has the potential to take advantage of each language to improve the performance of other languages, as the affect descriptors studied share features among languages. Furthermore, end-to-end deep learning architectures are able to recognize different affect classes, emotion and personality, automatically learning and processing the most relevant speech features. 

% \section*{Acknowledgment}


\ifCLASSOPTIONcaptionsoff
  \newpage
\fi


\bibliographystyle{IEEEtran}
\bibliography{bibliography}
%\begin{thebibliography}{1}


%\end{thebibliography}


% \end{IEEEbiography}



%\vfill


\end{document}